\documentclass{article}
\pdfoutput=1
\usepackage{collas/collas2022_conference,times}

\usepackage{amsmath,amsfonts,bm}

\def\eqref#1{equation~\ref{#1}}

\def\1{\bm{1}}

\def\vh{{\bm{h}}}

\def\vx{{\bm{x}}}

\def\mA{{\bm{A}}}
\def\mB{{\bm{B}}}
\def\mC{{\bm{C}}}

\def\mH{{\bm{H}}}
\def\mI{{\bm{I}}}
\def\mJ{{\bm{J}}}

\def\mL{{\bm{L}}}

\def\mN{{\bm{N}}}

\def\mV{{\bm{V}}}
\def\mW{{\bm{W}}}
\def\mX{{\bm{X}}}

\DeclareMathAlphabet{\mathsfit}{\encodingdefault}{\sfdefault}{m}{sl}
\SetMathAlphabet{\mathsfit}{bold}{\encodingdefault}{\sfdefault}{bx}{n}

\usepackage{algorithm}
\usepackage{algpseudocode}
\usepackage{subcaption}
\usepackage{multirow}
\usepackage{booktabs}
\usepackage{comment}
\usepackage{siunitx/siunitxx}
\usepackage[disable]{todonotes}
\usepackage[acronym]{glossaries}
\usepackage{enumitem}
\newacronym{cl}{CL}{continual learning}
\newacronym{ds}{DS}{dynamical systems}
\newacronym{rc}{RC}{reservoir computing}
\newacronym{vdp}{VdP}{Van-der-Pol}
\newacronym{l63}{L63}{Lorenz-63}
\newacronym{l96}{L96}{Lorenz-96}

\usepackage{hyperref}
\hypersetup{
    colorlinks=true,
    linkcolor=red,
    filecolor=magenta,
    urlcolor=blue,
    citecolor=purple,
    pdftitle={Overleaf Example},
    pdfpagemode=FullScreen,
    }

\title{Continual Learning of Dynamical Systems with Competitive Federated Reservoir Computing}

\author{Leonard Bereska\\
University of Amsterdam \\
\texttt{leonard.bereska@uva.nl} \\
\And 
Efstratios Gavves \\
University of Amsterdam \\
\texttt{efstratios.gavves@uva.nl} \\
}

\newcommand{\ours}{Ours} 

\collasfinalcopy

\begin{document}

\begin{center}
\maketitle
\end{center}


\begin{abstract}
Machine learning recently proved efficient in learning differential equations and dynamical systems from data.
However, the data is commonly assumed to originate from a single never-changing system.
In contrast, when modeling real-world dynamical processes, the data distribution often shifts due to changes in the underlying system dynamics.
\textit{Continual learning of these processes aims to rapidly adapt to abrupt system changes without forgetting previous dynamical regimes}.
This work proposes an approach to continual learning based on reservoir computing, a state-of-the-art method for training recurrent neural networks on complex spatiotemporal dynamical systems.
Reservoir computing fixes the recurrent network weights - hence these cannot be forgotten - and only updates linear projection \emph{heads} to the output.
We propose to train multiple competitive prediction heads concurrently.
Inspired by neuroscience's predictive coding, only the most predictive heads activate, laterally inhibiting and thus protecting the inactive heads from forgetting induced by interfering parameter updates.
We show that this multi-head reservoir minimizes interference and catastrophic forgetting on several dynamical systems, including the Van-der-Pol oscillator, the chaotic Lorenz attractor, and the high-dimensional Lorenz-96 weather model.
Our results suggest that reservoir computing is a promising candidate framework for the continual learning of dynamical systems.
We provide our code for data generation, method, and comparisons at \url{https://github.com/leonardbereska/multiheadreservoir}.
\end{abstract}
\glsresetall
\section{Introduction}
Modeling complex spatiotemporal data is a universal problem in diverse branches of science, such as climate- \citep{rolnick_tackling_2019, reichstein_deep_2019}, health- \citep{fresca_deep_2020} or neuroscience \citep{pandarinath_inferring_2018}. The complex systems in these domains are high-dimensional \textit{and} highly nonlinear, making it impossible to set up models from first principles \citep{strogatz_nonlinear_1994}.
As a result, recent years have seen data-driven learning algorithms for identifying nonlinear \gls{ds} become increasingly popular \citep{brunton_discovering_2016, pathak_model-free_2018, vlachas_data-driven_2018, lu_attractor_2018, raissi_multistep_2018, champion_data_2019, karniadakis_physics-informed_2021}.
However, these previous approaches to modeling complex data assume the data is independent and identically distributed (i.i.d.) and comes from a single, unchanging environment. In real-world settings, however, the generating distribution, as is described by the (partial) differential equations of the \gls{ds}, can shift abruptly. Thus, models trained with the i.i.d. assumption need to either continue training and risk overwriting previous knowledge or restart training from scratch if the distribution's change-point is known. To this end, we propose \gls{cl} of \gls{ds}, which will allow models to learn from and adapt to such novel and time-varying dynamical environments while remembering previously observed dynamics.

Recently, learning  \gls{ds} beyond the i.i.d. assumption from multiple environments has been proposed in the multi-task setting \citep{yin_leads_2021, wang_meta-learning_2021, kirchmeyer_generalizing_2022}, where data from different environments is available.
The more challenging and restrictive \gls{cl} setting assumes that access to data is only granted sequentially either due to security or privacy reasons or as a realistic assumption for a real-world intelligent agent.
Under such conditions, in contrast to biological agents, neural network-based models suffer from catastrophic forgetting \citep{mccloskey_catastrophic_1989, french_catastrophic_1999}, destroying previously learned knowledge as they train on novel data. This trade-off between
being stable enough not to forget the old and yet being plastic enough to rapidly learn the new is known as the \emph{stability-plasticity dilemma} \citep{mermillod_stability-plasticity_2013, grossberg_adaptive_2013, ditzler_learning_2015}.

So far, \gls{cl} research has focused heavily on minimizing forgetting on \emph{discriminative tasks} (e.g., classification) with \emph{static} data (e.g., images). However, neuroscientific evidence \citep{frolich_neuronal_2021} shows that memory is both \emph{generative} and \emph{sequential}, rather than \emph{discriminative} and \emph{static}. The generative nature of memory is evidenced by a neuroscientific theory called predictive coding, which suggests that the brain encodes top-down generative models to predict the next sensory input from lower levels \citep{buzsaki_space_2018}. Moreover, a tremendous amount of evidence from human ethology, physiology, and neuroscience shows that human memory operates via neuronal sequences as representations even in situations that seemingly provide only static sensory input \citep{frolich_neuronal_2021}. Despite this evidence highlighting the need for \gls{cl} in a generative sequential setting, research in this setting is scarce, with notable exceptions \citep{ororbia_continual_2020, cui_continuous_2016, grewal_going_2021, iyer_avoiding_2022}. Crucially, to the best of our knowledge, no studies have focused on applications to \gls{ds}.

To continually learn \gls{ds}, we propose \gls{rc} \citep{jaeger_echo_2001, maass_real-time_2002}, a state-of-the-art training scheme for recurrent neural networks on complex spatiotemporal \gls{ds}' data \citep{pathak_model-free_2018, lu_attractor_2018, rohm_model-free_2021}. In addition to promising good performance for \gls{ds} identification, the training mechanism promises to improve \gls{cl} by fixing the recurrent network weights, preventing catastrophic overwriting by subsequent parameter updates, and training only a linear output layer (\textit{head}). Similar ideas have been proposed for the static setting \citep{wortsman_supermasks_2020, ramanujan_whats_2020} and also for the sequential discriminative setting \citep{cossu_continual_2021-1, kobayashi_continual_2019}.
In this work, we draw inspiration from predictive coding to protect the reservoir output \textit{head} from interference \citep{ororbia_lifelong_2021}. Accordingly, we train multiple heads concurrently, each trying to predict the observed sequence. The best matching prediction heads activate sparsely and thereby laterally inhibit competing ones \citep{aljundi_selfless_2018}\todo{cite yuguo}. Only the active heads are updated in a continual manner with a federated \gls{rc} \citep{bacciu_federated_2021} update scheme. Thus, the inactive heads' knowledge of previous environments stays untouched. As a result, we resolve the plasticity-stability dilemma by being simultaneously highly stable (fixed weights) and highly plastic (new head for new environment).

In summary, our contributions are:
	\begin{enumerate}[label=\emph{\roman*).}]
	    \item We propose competitive multi-head reservoir computing as a brain-inspired framework to continually train with reservoir computing.
	    \item We are, to the best of our knowledge, the first to address continual learning for dynamical systems.
	    \item Thereby, we provide a use case for reservoir computing for the understudied setting of Sequential Generative continual learning more generally.
	\end{enumerate}
We show the effectiveness of this competitive multi-head reservoir in minimizing forgetting on several benchmark \gls{ds}, namely on the Van-der-Pol oscillator, the chaotic Lorenz attractor, and the high-dimensional Lorenz-96 atmosphere model.

\section{Method}\label{sec:method}

In the following, we describe our method for training a multi-head reservoir to continuously adapt to different dynamical systems environments.
\paragraph{Desiderata for continually generating sequences.}
We start by listing several properties that we consider desirable for a sequential generative continual learning algorithm:
	\begin{enumerate}[label=\Roman*.]
		\item \emph{Multiple simultaneous predictions} \citep{hawkins_why_2016}: For a given context, there are often multiple possible future trajectories. It is crucial to keep multiple competing hypotheses simultaneously and evaluate the likelihood of each hypothesis, especially when the context is ambiguous.
		\item \emph{Continual training and testing} \citep{hawkins_why_2016}: The difference between the training and testing set should be minimal to allow online adjustment to change.
		\item \emph{No external task supervision} \citep{lesort_continual_2020}: In real-world settings, the model should determine the current task from the data alone without supervision in the form of task labels.
		\item \emph{No storage of raw sensory data} \citep{lesort_continual_2020}: Our goal is to develop a sequence memory that compresses high-dimensional perceptual data and extracts information for prediction. Storing raw sensory data violates that objective.
	\end{enumerate}

	\begin{figure}[!htp]
		\includegraphics[width=.99\linewidth]{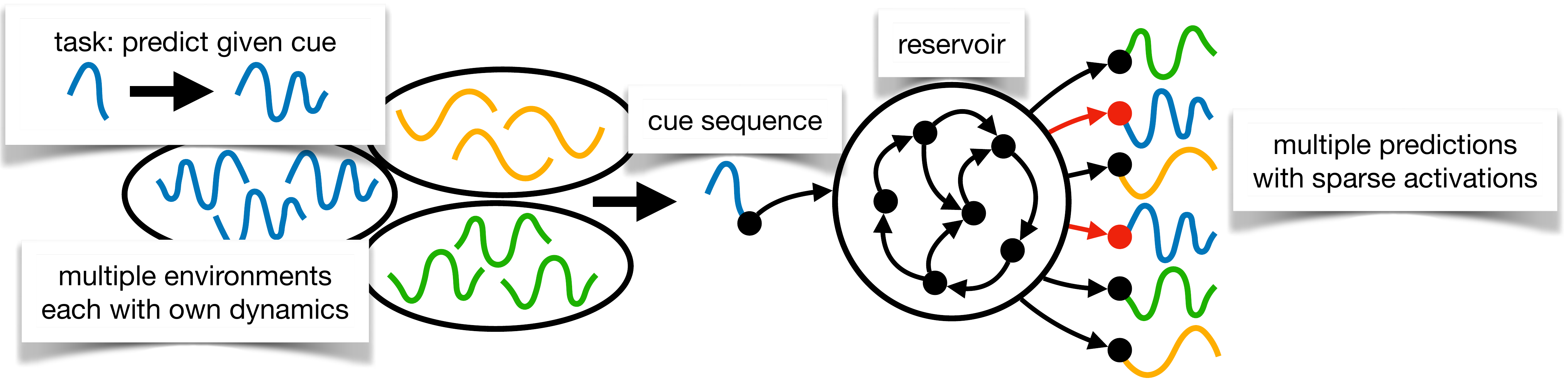}
		\caption{Overview of task setting and method: The model is trained on sequences drawn from multiple environments. Each environment is encountered only once, without replay. Given a fragment of a training sequence, the model should associate/predict the full training sequence. Since it is trained successively on the environments, the challenge is not to forget dynamics from previous environments when encountering new ones. A fixed-weight reservoir with multiple prediction heads of which only a subset sparsely activates and is updated achieves this.}
		\label{fig:method}
	\end{figure}

	\paragraph{Sequential Generative Continual Learning.}\label{sec:seqgen}
		The learning algorithm is trained on sequences $\mX_{1:T} = \{\vx_t| t=1,\ldots,T\}$ of $d$-dimensional vectors $\vx_t\in \mathbb{R}^d$. First, sequences stem from the same distribution, and at some point (unknown to the algorithm), the distribution shifts abruptly to a different environment, \textit{e.g.,} other underlying system equations.
		Given a sequence fragment $\vx_{1:\tau}$ as cue, the task is to remember the future trajectory of that sequence $\vx_{\tau:T}$.
		The challenge is to not forget sequences from the previous environment after training only on sequences from the novel environment.
		Accordingly, the learning algorithm is trained on $K$ environments $\{\mathcal{E}_k\}_{k=1,..., K}$ in succession, the number of which is unknown to the algorithm.
		The training data as a whole is thus an ordered collection of sequences $\{\mX_{1:\tau}^{(n)}\}_{n=1,\ldots,N}$, where the algorithm is agnostic to the origin-environment of each sequence $\mX_{1:\tau}^{(n)}$.

	\paragraph{Reservoir Computing}
		starts with a randomly connected recurrent neural network. By fixing the recurrent network weights, \gls{rc} forgoes the notorious vanishing gradient problem arising from backpropagation through time.
		The state of the recurrent neural network of $M$ neurons is given by its hidden activations $\vh \in \mathbb{R}^{M}$, which are connected by \textit{randomly initialized and fixed} matrix $\mW_{\mathrm{h}}$.
		An input sequence $\mX_{1:\tau}$ is embedded by a linear map $\mW_{\mathrm{i}}$ to the state $\vh_\tau$:
		\begin{align}\label{eq:reservoir}
		    \begin{split}
			    \vh_1 &= \sigma (\mW_{\mathrm{h}} \vh_0 +\mW_{\mathrm{i}}\vx_1) \\
			\vdots &\phantom{a}\hspace{1cm} \vdots\\
			\vh_\tau &= \sigma (\mW_{\mathrm{h}} \vh_{\tau -1} + \mW_{\mathrm{i}} \vx_\tau).
		    \end{split}
		\end{align}
		The reservoir linearly predicts the next input $\vx_{\tau +1}$:
		\begin{equation}
			\hat \vx_{\tau +1} = \mW_{\mathrm{o}} \vh_\tau,
		\end{equation}
		with $\mW_{\mathrm{o}}$ mapping from hidden activation to output.
		Only the parameters of this output matrix $\mW_{\mathrm{o}}$ are trained. Hence, the optimal values can be obtained analytically via ridge regression (with regularization parameter $\lambda$) as:
		\begin{equation}\label{eq:regression}
			\mW_{\mathrm{o}} = (\mH^T\mH + \lambda \mI)^{-1} \mH^T\mX,
		\end{equation}
		where $\mH$ are the hidden states, $\mI$ the identity matrix, and $\mX$ the targets of the regression.

	\paragraph{Federated Reservoir Computing.}
		When training is decentralized onto multiple clients for data security or privacy reasons, federated learning provides a framework to integrate the learning progress of these clients into one shared machine learning model.
		In our case, the training data is not scattered in space but in time: continual learning prohibits accessing all data points at once. Therefore, we propose transferring the machinery of federated learning to continual learning by interpreting different clients as different points in time.
		Following \citet{bacciu_federated_2021}, we break up the analytic (and thus instantaneous) calculation of the optimal output weights in Eq. \ref{eq:regression} into components:
		\begin{align}\label{eq:defineAB}
			\begin{split}
			\mA&=\mH^T\mH \\
			\mB&=\mH^T\mX,
			\end{split}
		\end{align}
		such that Eq. \ref{eq:regression} turns into $\mW_{\mathrm{o}} = (\mA+\lambda \mI)^{-1} \mB$.
		These terms $\mA$ and $\mB$ are then additive \textit{w.r.t.} successively incoming data, as multiple iterative data fragments $\mH = \begin{pmatrix}\vh_{1} \\ \vh_{2} \end{pmatrix}$,  $\mX = \begin{pmatrix}\vx_{1} \\ \vx_{2} \end{pmatrix}$ decompose into sums:  $\mA = \vh_1^T \vh_1 + \vh_2^T \vh_2$ and $\mB = \vh_1^T \vx_1 + \vh_2^T \vx_2$.
		Hence, for training successively at time step $t$, it suffices to keep a copy of $\mA^{(t-1)}$ and $\mB^{(t-1)}$:
			\begin{align}\label{eq:updateAB}
			\begin{split}
			\mA^{(t)} &= \mA^{(t-1)} + \tilde \mA \\
			\mB^{(t)} &= \mB^{(t-1)} + \tilde \mB,
			\end{split}
			\end{align}
		where  $\tilde \mA$,  $\tilde \mB$ are calculated on incoming data at time step $t$ according to definition in Eq. \ref{eq:defineAB}. Output weights are again calculated as:
			\begin{equation}\label{eq:update_head}
				\mW_{\mathrm{o}}^{(t)} = (\mA^{(t)}+\lambda \mI)^{-1} \mB^{(t)}.
			\end{equation}

\paragraph{Competitive Multi-Head Reservoir.}
	We want the learning algorithm to make multiple predictions when presented with a sequence (desideratum 1.).
	Hence, we train a reservoir with $L$ linear heads. The number of prediction heads $L$ should be much higher than the number of environments $K$ that the algorithm is expected to learn continually: $L\gg K$. Although $K$ is assumed unknown, in principle, we can dynamically add (with linearly rising memory cost) new heads when shifts in environments are detected.\\
	To iteratively learn with federated reservoir computing, we then simultaneously maintain $L$ prediction heads $\{\mW_o^{(l)}| l = 1, \ldots,  L\}$, corresponding to $L$ respective matrices $\mA^{(l)}, \mB^{(l)}$.
	All heads compete for predicting the sequence most accurately, as measured by the Mean Squared Error (MSE). Given a sequence $\mX_{1:T}$, the reservoir state $\vh_t$ for time step $t$ is obtained by embedding $\mX_{1:t}$ via Eq. \ref{eq:reservoir}. The Squared Error (SE) for head $l$ is then:
	\begin{equation}
		\textrm{SE}^{(l)} = (\hat\vx_{t+1}^{(l)} -  \vx_{t+1})^2 = (\mW_{o}^{(l)} \vh_t - \vx_{t+1})^2.
	\end{equation}
	For each sequence we take the average over all time steps $t$ to obtain $\textrm{MSE}^{(l)}$ from $\textrm{SE}^{(l)}$.
	Heads with the lowest MSE activate unless a new environment is detected, in which case new (previously never active) heads become activated.
	Only active heads receive parameter updates, which protects inactive heads from catastrophic forgetting. We visualize the competitive multi-head reservoir and its application to continual learning of dynamical systems in Fig. \ref{fig:method}.
	The complete algorithm in pseudocode is revealed in Algo. \ref{algorithm}.
	\begin{algorithm}[!htp]\label{algorithm}
		\caption{Competitive Multi-Head Reservoir Computing}\label{algorithm}
		\begin{algorithmic}
			\Require Ordered collection of input sequences $\{\mX_{1:\tau}^{(n)}\}_{n=1,\ldots,N}$.
			\Require Novel environment detection threshold $\theta_{\textrm{new env}}$.
			\State Initialize (sparse-random) reservoir $\mW_h, \mW_i$.
			\State Initialize (randomly) $L$ heads $\{\mW_{o}^{(l)}\}_{l=1,\ldots,L}$.
			\State Initialize $\mA^{(l)}, \mB^{(l)}$ to zero (matrices).
			\State Initialize \textit{never-before-active heads} as set that contains all heads.
			\State Initialize \textit{previous active heads} as empty set.
			\For{$n = 1$ to $N$}
			\State Embed sequence $\mX_{1:\tau}^{(n)}$ according to Eq. \ref{eq:reservoir} to obtain reservoir state $\vh_\tau$.
			\State Calculate ahead prediction $\textrm{MSE}^{(l)}$ for each head $\mW_{o}^{(l)}$.
			\State
			\If{detect novel environment with $\textrm{MSE}^{(l)}/\textrm{MSE}_{\textrm{last}}^{(l)}>\theta_{\textrm{new env}}$ for any $l$ of \textit{previous active heads}}
				\State \textit{active heads} $\gets$ random heads from \textit{never-before-active heads}.
			\Else
			\State \textit{active heads} $\gets$ heads with lowest MSE.
			\EndIf
			\State
			\State Update $\mA^{(l)}, \mB^{(l)}$ and $\mW_{o}^{(l)}$ for $l$ in \textit{active heads} according to Eq. \ref{eq:updateAB} and \ref{eq:update_head}.
			\State
			\State $\textrm{MSE}_{\textrm{last}}^{(l)} \gets \textrm{MSE}^{(l)}$, for all $l$.
			\State \textit{previous active heads} $\gets$ \textit{active heads}.
			\State Remove \textit{active heads} from \textit{never-before-active heads}.
			\EndFor
		\end{algorithmic}
	\end{algorithm}

\section{Experiments}
	We experiment on three families of dynamical systems that exhibit a diverse set of dynamical regimes: The nonlinear \gls{vdp} oscillator is a representative of limit cycle behavior, the famous chaotic \gls{l63} attractor is a simple chaotic system and the \gls{l96} atmosphere model is a high-dimensional chaotic system.
	We instantiate the respective differential equations with varying sets of parameter values to obtain four different environments per system.
	Thus, different environments share the same functional form of the underlying equations but differ in particular values for the parameters. \textit{Therefore, a shift from one environment to the next can be interpreted as an abrupt bifurcation}.

	\subsection{Datasets and environments}

		\begin{figure}[!htp]
		\centering
		\begin{subfigure}{0.32\textwidth}
		    \includegraphics[height=0.85\textwidth]{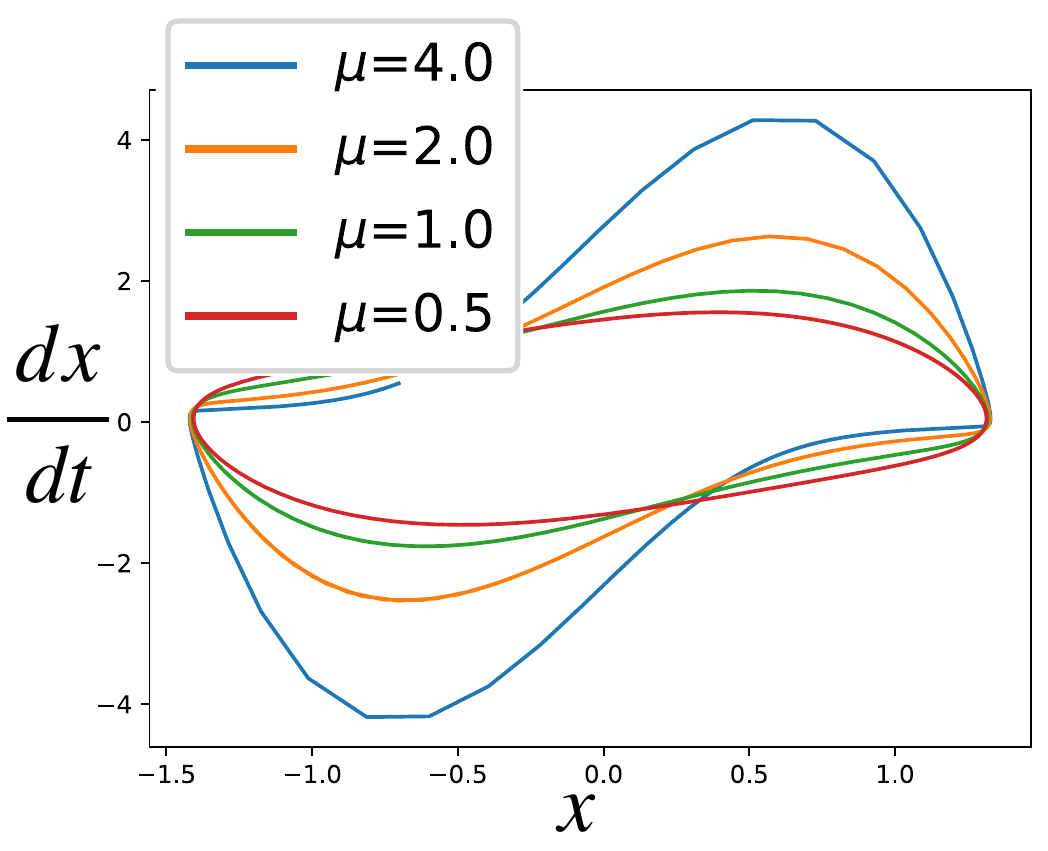}
		    \caption{}
		    \label{fig:vdp}
		\end{subfigure}
		\hfill
		\begin{subfigure}{0.32\textwidth}
		    \includegraphics[height=0.85\textwidth]{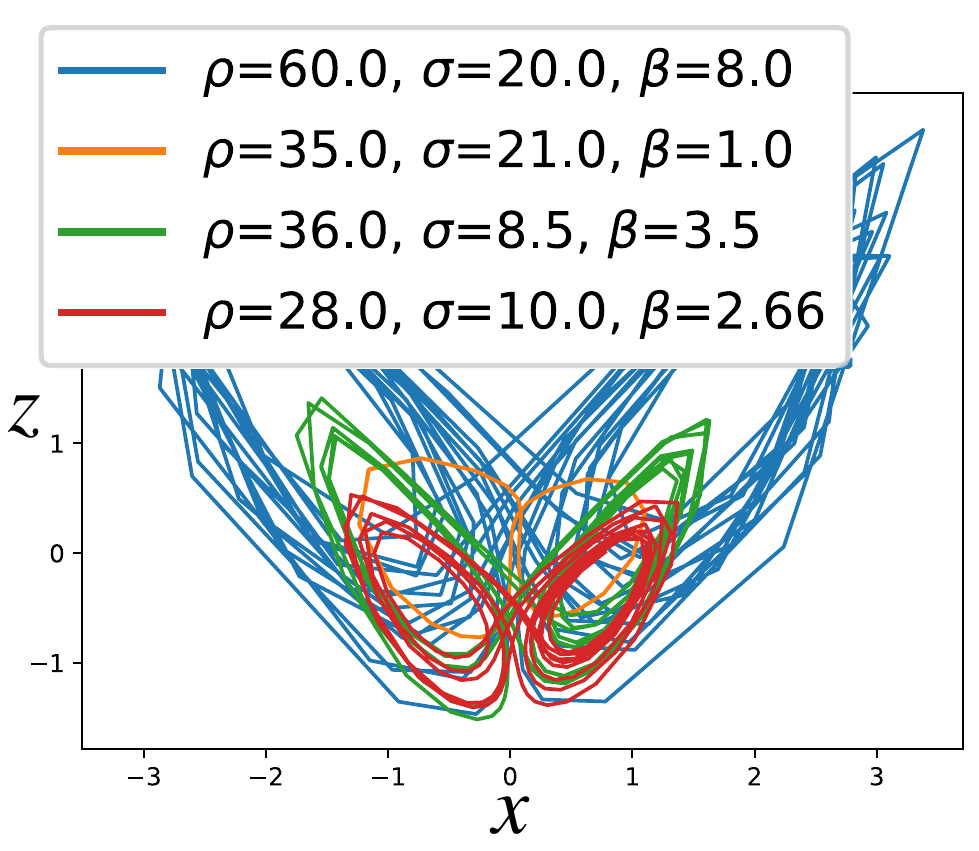}
		    \caption{}
		    \label{fig:l63}
		\end{subfigure}
		\hfill
		\begin{subfigure}{0.32\textwidth}
		    \includegraphics[height=0.85\textwidth]{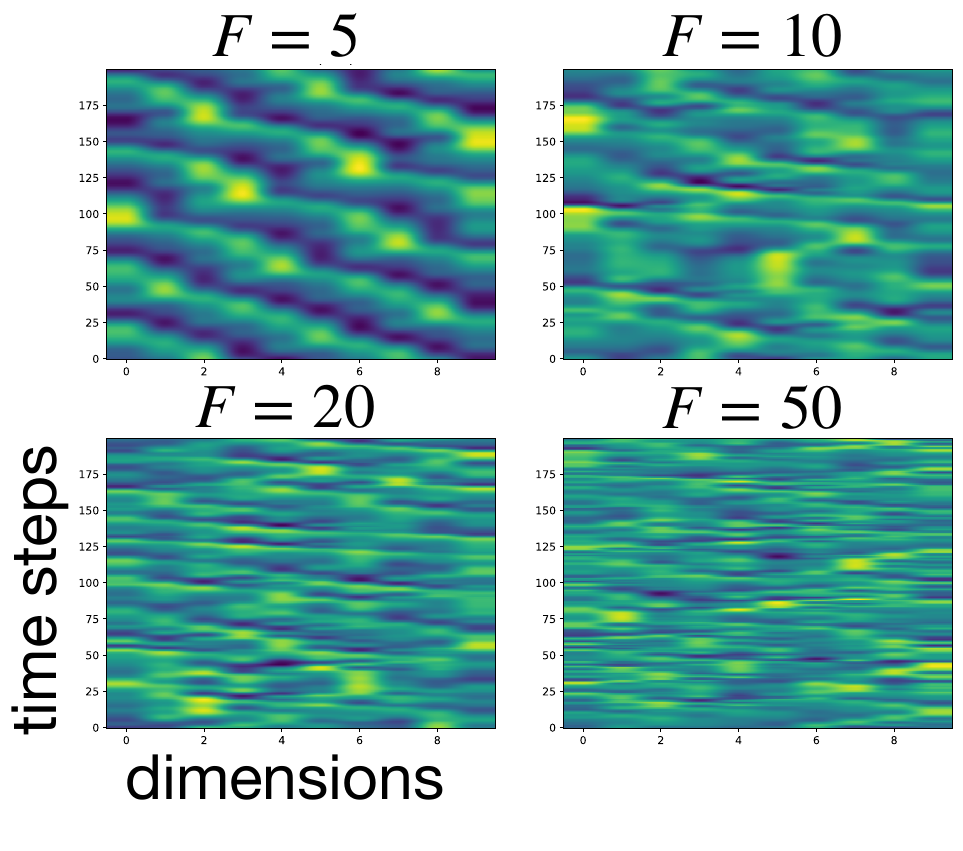}
		    \caption{}
		    \label{fig:l96}
		\end{subfigure}

			\caption{Environments for the (a) Van-der-Pol oscillator, the (b) Lorenz-63 attractor, and the (c) Lorenz-96 model.}
		\label{fig:envs}
		\end{figure}

		\glsreset{vdp}
		\glsreset{l63}
		\glsreset{l96}
			\paragraph{\gls{vdp}.}
				The \gls{vdp} oscillator is a non-conservative oscillator with nonlinear damping. It is defined through the second-order differential equation:
				\begin{equation}
				\frac{\mathrm{d}^2x}{\mathrm{d}t^2}-\mu (1-x^2) \frac{\mathrm{d}x}{\mathrm{d}t}+x = 0,
				\end{equation}
				and evolves on a limit cycle. The damping parameter $\mu$ characterizes the shape of this cycle. Lower values of $\mu$ result in a more circular shape; for $\mu=0$ (no damping), the system becomes the conservative simple harmonic oscillator. Our environments defined by $\mu \in \{0.5, 1, 2, 4\}$ are shown in Fig. \ref{fig:vdp}.

			\paragraph{\gls{l63}.}
				The classical three-dimensional Lorenz attractor \citep{lorenz_deterministic_1963} was the first example of a low-dimensional system with chaotic solutions. The \gls{l63} system is described by the differential equations:
				\begin{equation}
						\frac{\mathrm{d}x}{\mathrm{d}t} = \sigma (y - x),
						\frac{\mathrm{d}y}{\mathrm{d}t} = x (\rho - z) - y,
						\frac{\mathrm{d}z}{\mathrm{d}t} = x y - \beta z.
				\end{equation}
				It is frequently used as a benchmark system for \gls{ds} reconstruction algorithms \citep{pathak_model-free_2018}. We use the following parameter instantiations as environments: $(\rho, \sigma, \beta) \in \{(28, 10, 2.66), (36, 8.5, 3.5), (35, 21, 1), (60, 20, 8)\}$ (visualized in Fig. \ref{fig:l63}).

			\paragraph{\gls{l96}.}
				The Lorenz-96 model \citep{lorenz_predictability_1996} was originally constructed by Edward Lorenz as a model for the laminar nature of the earth's atmosphere. The dimensionality $D$ can be freely chosen, as the system is recursively defined, by
				\begin{equation}\label{eq:l96}
				\frac{\mathrm{d}x_i}{\mathrm{d}t} = (x_{i+1}-x_{i-2})x_{i-1} - x_i + F, \ i=1, \dots, D;
				\end{equation}
				with the forcing term $F$ as only parameter, defining the environments by $F \in \{5, 10, 20, 50\}$ (Fig. \ref{fig:l96}). It is characterized by nearest-neighbor interactions ($x_i$ interacting with neighboring $x_{i-2}, x_{i-1},$ and $x_{i+1}$, as directly seen from Eq. \ref{eq:l96} ) and is thus easily visualized as a 1D image.\todo{refer to figure, reconstruction plot} Like the \gls{l63}, it is a common benchmark for \gls{ds} reconstruction.

			\paragraph{Simulation details.}
				We simulate the systems with $\texttt{scipy.odeint}$, with step size $0.05$. The initial state is randomly (uniform distribution for each dimension) assigned. To only obtain the limit distribution, we cut transient behavior by dropping the first $100$ time steps. We simulate $10$ sequences per environment, each for $200$ time steps. And split the data into $7$ sequences for training and $3$ sequences for testing.
				We standardize the data (subtract mean and divide by standard deviation) jointly over all four environments per system.

	\subsection{Results}
	We consider the \gls{vdp}, \gls{l63}, and \gls{l96} datasets, where we train on $4$ environments with $7$ trajectories per environment.
	For evaluation, we create a separate test set of $3$ trajectories, where the multi-head reservoir's active head is chosen according to the $10$-time step cue (out of a total of $200$ time steps per sequence).
	To quantify the performance, we choose the one-step-ahead prediction MSE (in units of $10^{-3}$), averaging over $10$ experiments for each setting.
	The MSE is given in absolute terms, but as the data are normalized, they can also be interpreted as relative to the standard deviation of the data.

	\paragraph{Comparing to methods for continual learning.}
		For lack of comparison methods in the domain of generative sequential continual learning, we compare our method, the competitive multi-head reservoir, to two constructed baselines that unfortunately do not satisfy our desiderata completely:
		\begin{enumerate}[label=\emph{\roman*).}]

			\item Similarly to \citet{cossu_continual_2021-1}, we combine a Long Short-Term Memory (LSTM) \citep{hochreiter_long_1997} model with a general-purpose continual learning regularization technique, namely Elastic Weight Consolidation (EWC) \citep{kirkpatrick_overcoming_2017}.
			This method does not make multiple simultaneous predictions (contrary to our first desideratum, Sec. \ref{sec:method}) and needs to be supplied with task labels (disobeying our third desideratum).
			\item In addition, we compare to a simple replay baseline again using a simple LSTM. Replay is considered one of the most effective continual learning methods \citep{van_de_ven_brain-inspired_2020} if one allows for storing raw data (we do not consider that desirable: see our fourth desideratum). Following \citet{lopez-paz_gradient_2017} we save the last sequence of each dataset in a buffer and randomly intersperse the regular training with buffer training steps with a frequency of 1\%.
		\end{enumerate}
		The comparison to these baselines is shown in Tab. \ref{tab:comparison_continual}.

\begin{table}[!htp]
	\caption{Comparing to other continual learning methods. We give the one-step-ahead prediction mean squared error (in units of $10^{-3}$), and the standard error of the mean in brackets.
	Note that the fourth environment is not indicative of continual learning performance as it is trained last and thus not subject to catastrophic forgetting.
	}
\begin{center}
\begin{tabular}{ccc cccc}

\toprule
	\multirow{2}*{Dataset}& \multirow{2}*{Desiderata}& \multirow{2}*{Method}& \multicolumn{4}{c}{Environment}  \\
	 & & & 1 & 2 & 3 & 4  \\
\midrule
	\multirow{3}*{\gls{vdp}}& II, IV& LSTM+EWC& \num{10.1+-0.16} & \num{7.87+-0.15} & \num{4.3+-0.11} & \num{0.000306+-0.0000082} \\
	& II & LSTM+Replay& \num{0.5419999999999999+-0.013} & \num{0.28600000000000003+-0.0084} & \num{0.589+-0.016} & \num{5.88+-0.27999999999999997} \\
	& I, II, III, IV& \ours & \num{1.2979999999999998+-0.13999999999999999} & \num{1.7650000000000001+-0.19} & \num{3.478+-0.41} & \num{20.490000000000002+-4.6} \\
\midrule
	\multirow{3}*{\gls{l63}}& II, IV & LSTM+EWC& \num{463.0+-6.3} & \num{424.0+-5.1000000000000005} & \num{238.0+-3.3} & \num{0.053399999999999996+-0.0026} \\
	& II &  LSTM+Replay& \num{19.400000000000002+-0.22} & \num{19.7+-0.25} & \num{9.0+-0.13999999999999999} & \num{454.0+-5.1000000000000005} \\
	 & I, II, III, IV & \ours & \num{7.749+-0.7699999999999999} & \num{60.26+-3.8} & \num{3.595+-0.3} & \num{26.03+-1.8} \\
\midrule
	\multirow{3}*{\gls{l96}}& II, IV& LSTM+EWC& \num{162.0+-0.6} & \num{133.0+-0.53} & \num{111.0+-0.61} & \num{332.0+-2.9} \\
	& II &  LSTM+Replay& \num{1.7799999999999998+-0.018000000000000002} & \num{5.44+-0.043000000000000003} & \num{27.7+-0.22} & \num{430.0+-3.5} \\
	&  I, II, III, IV &\ours & \num{2.932+-0.056} & \num{15.94+-0.34} & \num{197.0+-6.7} & \num{2473.0+-56.0} \\
\bottomrule
\end{tabular}
\end{center}
\label{tab:comparison_continual}
\end{table}

	\paragraph{Comparing the continual, single-task and multi-task setting.}
		To provide some context, we also compare to two relaxations of continual learning, namely the:
		\begin{enumerate}[label=\emph{\roman*).}]
			\item \textit{Single-task} setting: Training and evaluating each environment separately. This constitutes an upper bound on performance (\textit{i.e.} lower bound on MSE).
			\item \textit{Multi-task} setting: Joint training, with access to all environments at once. While accessing all data at once may benefit training an LSTM, for the RC training, this setting can be seen as a lower bound on performance, as a single head needs to cater to different environments and is thus forced to interpolate across environments with merely a linear layer.
		\end{enumerate}
		Note that replay can be seen as an interpolation between the continual and the multi-task setting, but we still list it under the continual learning methods in Tab. \ref{tab:comparison_continual}.
		For these non-continual settings, we compare to RC, a standard LSTM, and Sparse Identification of Nonlinear Dynamical systems (SINDy) \citep{brunton_discovering_2016}, a state-of-the-art method for identification of dynamical system equations.
		SINDy sparsely regresses coefficients for a library of terms and hence directly learns the governing equations of the dynamical system.
		We use the publicly available \texttt{PySINDy} package \citep{de_silva_pysindy_2020} with default parameters (which are chosen to work very well on the popular DS benchmarks that we work with by default).

		\subsection{Discussion}

		\paragraph{Continual learning.}
			The LSTM seems to be the more powerful approximation method in comparison to RC, as becomes clear from the orders-of-magnitudes-better performance in the single- and multi-task settings, cf. Tab. \ref{tab:comparison_setting}.
			Despite being generally more powerful, in the continual learning setting, the LSTM forgets previous environments catastrophically, and the regularization by elastic weight consolidation does not prevent this.
			If trained with replay, on the other hand, the LSTM does not suffer from catastrophic forgetting, as can be expected from interpolation between the continual and the multi-task setting.
			In fact, it performs on par with the multi-task setting (cf. Tab. \ref{tab:comparison_setting}).
			Comparing our competitive multi-head \gls{rc} to these two methods shows that catastrophic forgetting is indeed mitigated.
			As shown in Tab. \ref{tab:comparison_continual}, our competitive multi-head reservoir remembers how to predict previous environments in the continual setting.

		\paragraph{Single- and multi-task setting.}
			The LSTM performs well both in the single-task and the multi-task setting.
			In contrast, for \gls{rc}, multi-tasking overstretches the capacity of a single prediction head. Note that the competitive multi-head reservoir keeps the performance close to its upper bound, the single-task \gls{rc}.
			SINDy works best in the single-task setting (for which it is designed) and naturally struggles with integrating multiple environments into one dynamical system equation.

		\begin{table}[!htp]
	\caption{Comparing to single- and multi-task settings. We give the one-step-ahead prediction mean squared error (in units of $10^{-3}$), and the standard error of the mean in brackets.}
\begin{center}
\begin{tabular}{ccc cccc}

\toprule
	\multirow{2}*{Dataset}& \multirow{2}*{Setting}& \multirow{2}*{Method}& \multicolumn{4}{c}{Environment}  \\
	 & & & 1 & 2 & 3 & 4  \\
\midrule
	\multirow{9}*{\gls{vdp}} & \multirow{3}*{Single-task} & RC & \num{4.45 \pm 0.23} & \num{2.80 \pm 0.25}  &\num{8.545 \pm 0.71}   &  \num{10.34 \pm 0.99}\\
	&  & LSTM  & \num{0.0195+-0.00028000000000000003} & \num{0.032799999999999996+-0.00043} & \num{0.0554+-0.00063} & \num{0.0539+-0.002} \\
	&  & SINDy & \num{0.273 \pm 0.0044} & \num{1.18 \pm 0.043}  &\num{5.03 \pm 0.15}   &  \num{19.0 \pm 0.83}\\
	\cmidrule{2-7}
	& \multirow{3}*{Multi-task} & RC &\num{20.26+-2.2} & \num{15.15+-1.4} & \num{3.2880000000000003+-0.33} & \num{7.3709999999999996+-1.4} \\
	& &LSTM& \num{0.40900000000000003+-0.0070999999999999995} & \num{0.162+-0.0035} & \num{0.372+-0.0092} & \num{1.23+-0.049} \\
	&  & SINDy & \num{0.451 \pm 0.011} & \num{1.28 \pm 0.025}  &\num{5.56 \pm 0.14}   &  \num{19.4 \pm 0.86 }\\
	\cmidrule{2-7}
	& \multirow{1}*{Continual} & \ours & \num{1.2979999999999998+-0.13999999999999999} & \num{1.7650000000000001+-0.19} & \num{3.478+-0.41} & \num{20.490000000000002+-4.6} \\
\midrule
	\multirow{9}*{\gls{l63}}& \multirow{3}*{Single-task} & RC  & \num{8.503 \pm 0.91}& \num{31.02 \pm 2.2}& \num{4.005 \pm 0.38}  & \num{24.47 \pm 1.5} \\
	&  &LSTM  & \num{0.035800000000000005+-0.00089} & \num{0.0553+-0.00082} & \num{0.00294+-0.000034} & \num{0.261+-0.0052} \\
	&  & SINDy & \num{0.385 \pm 0.0066} & \num{2.33 \pm 0.035}  &\num{0.354 \pm 0.0072}   &  \num{262 \pm 3.5}\\
	\cmidrule{2-7}
	& \multirow{3}*{Multi-task}& RC& \num{7.4910000000000005+-0.33} & \num{33.92+-2.3} & \num{4.707+-0.25999999999999995} & \num{180.4+-12.0} \\
	& & LSTM& \num{9.84+-0.2} & \num{40.599999999999994+-0.69} & \num{19.900000000000002+-0.27999999999999997} & \num{101.0+-1.5} \\
	&  & SINDy & \num{34.7 \pm 0.89} & \num{60.9 \pm 1.3}  & \num{35.7 \pm 0.62}  &  diverging \\
	\cmidrule{2-7}
	& \multirow{1}*{Continual}& \ours & \num{7.749+-0.7699999999999999} & \num{60.26+-3.8} & \num{3.595+-0.3} & \num{26.03+-1.8} \\
\midrule
	\multirow{9}*{\gls{l96}}& \multirow{3}*{Single-task}& RC  & \num{0.9859 \pm 0.025} & \num{7.204 \pm 0.18}&  \num{189.300 \pm 4.100} & \num{4181.000 \pm 81.000} \\
	&  & LSTM  & \num{0.14200000000000002+-0.0011} & \num{1.39+-0.01} & \num{20.400000000000002+-0.17} & \num{764.0+-5.4} \\
	&  & SINDy & \num{0.0239 \pm 0.00037} & \num{0.523 \pm 0.0046}  &\num{14.2 \pm 0.12}   &  \num{783 \pm 9}\\
	\cmidrule{2-7}
	& \multirow{3}*{Multi-task}&RC& \num{70.71+-1.5} & \num{202.5+-3.9} & \num{651.1+-13.0} & \num{2581.0+-56.0} \\
	& & LSTM& \num{0.359+-0.0022} & \num{1.49+-0.01} & \num{9.299999999999999+-0.068} & \num{169.0+-1.9} \\
	&  & SINDy & \num{41.0 \pm 0.32} & \num{36.1 \pm 0.3}  &\num{44.9 \pm 0.44}   &  \num{404 \pm 4.9}\\
	\cmidrule{2-7}
	& \multirow{1}*{Continual}&\ours & \num{2.932+-0.056} & \num{15.94+-0.34} & \num{197.0+-6.7} & \num{2473.0+-56.0} \\
\bottomrule
\end{tabular}
\end{center}
\label{tab:comparison_setting}
\end{table}

	\subsection{Experimental details}
		For initializing the reservoir, we use a radius of ${0.6}$, sparsity of ${0.01}$, and a reservoir size of $M ={1000}$, which are default hyperparameters we take from \citet{pathak_model-free_2018}.
		We take the first $10$ time steps of each sequence as a cue for initializing the hidden state, and when predicting the future (on test data), the first $5$ steps, while the following $5$ steps are used for determining the prediction head.
		We obtain the regularization parameter $\lambda$ for the reservoir by validation on a separate validation set (cf. Appendix, Fig. \ref{fig:regularization}), resulting in $\lambda_{\textrm{\gls{l63}}}={10^{-6}}$, $\lambda_{\textrm{\gls{l96}}}={5}$, and $\lambda_{\textrm{\gls{vdp}}}= {10^{-6}}$. We use the same hyperparameters for all settings.
		For the multi-head architecture, we use ten heads, with one head active at any given time. We determine the drift detection threshold for detecting a new environment on the validation set (cf. Appendix, Fig. \ref{fig:threshold}), where we exclude values below $2$ due to too many false-positive detections.
		In practice, an order (or even multiple orders) of magnitude increase in MSE is typical as the environment changes. Hence, the drift detection works for an extensive range of threshold values between $2$ and $100$, as an analysis (on a separate validation set) of the accuracy shows (cf. Appendix, Fig. \ref{fig:accuracy}). We choose a doubling in error (factor $2$ increase, thus a value of $2$ for the threshold) as a drift detection threshold.
		We run all experiments on an Apple M1. Running all experiments related to \gls{rc} (three datasets, three methods, four environments, hence 36 settings (as in Tab. \ref{tab:comparison_setting} with 10 data points per setting) takes only a couple of minutes, in contrast to the LSTM training that takes a couple of hours.
		This shows the computational efficiency of the one-shot linear regressions used by RC in comparison to the gradient-based LSTM training.

\section{Related Work}

\subsection{Multi-Environment Dynamical Systems}

Most data-driven approaches to identify differential equations \citep{pathak_model-free_2018, vlachas_data-driven_2018, brunton_discovering_2016, lu_attractor_2018, raissi_multistep_2018, champion_data_2019, chang_reviewing_2021, li_fourier_2020, chen_neural_2019} assume the data to be i.i.d., \textit{i.e.,} from one (single-environment) dynamical system.
To integrate multiple dynamical environments into one shared model, several works \citep{yin_leads_2021, kirchmeyer_generalizing_2022, wang_meta-learning_2021} propose to learn from several dynamical systems in a multi-task setting. These methods generalize over environments by deconstructing the learned function into a general and an environment-specific term. In contrast, our setting for continual learning prohibits access to all data at once.
Another line of work considers sequential access to data with abrupt distributional shifts \citep{quade_sparse_2018, li_rapid_2020}. While these methods focus on adaptation to the novel distribution, they do not remember the system before the shift, in contrast to our method. To date, the \gls{cl} setting has not been considered for the case of \gls{ds}.

\subsection{Continual Learning Beyond the Stability-Plasticity Dilemma}\label{ssec:noninterference}
As described by the \emph{stability-plasticity dilemma}, \gls{cl} is characterized by a trade-off between being stable enough not to forget the old and yet being plastic enough to rapidly learn the new. Attempts to continually integrate new observations without interfering with existing representations can be grouped into three strands \citep{parisi_continual_2019}:
	\textit{i)} Replay of previous data needs, relaxing the memory constraint (from our desiderata in Sec. \ref{sec:method}) to store a buffer of representative data points \citep{sprechmann_memory-based_2018, aljundi_task-free_2019} or training a generative model from which to sample previous experience \citep{shin_continual_2017, kemker_fearnet_2018, ostapenko_learning_2019, wu_memory_2018, van_de_ven_brain-inspired_2020, rolnick_experience_2019}.
	\textit{ii)} Expanding the model capacity to prevent overwriting previous memory. \citep{rusu_progressive_2016, yoon_lifelong_2018, coop_ensemble_2013}
	\textit{iii)} Ensuring that subsequent training does not interfere by regularizing task-relevant weights \citep{kirkpatrick_overcoming_2017, zenke_continual_2017, aljundi_memory_2018}, or isolating parameters of different tasks into orthogonal subspaces \citep{duncker_organizing_2020}, subnetworks \citep{masse_alleviating_2018, von_oswald_learning_2021}, submasks \citep{mallya_packnet_2018, mallya_piggyback_2018, serra_overcoming_2018,  ramanujan_whats_2020, fernando_pathnet_2017} or sparse representations \citep{ahmad_how_2016, ororbia_continual_2020, wortsman_supermasks_2020, iyer_avoiding_2022}.

While the plastic model components must be non-interfering for different tasks, any component shared among tasks minimizes forgetting by maximizing stability. This consideration leads to the idea of fixing weights in advance \citep{frankle_lottery_2019, ramanujan_whats_2020, wortsman_supermasks_2020}. For sequential data, fixing recurrent network weights is known as Reservoir Computing (\gls{rc}) \citep{jaeger_echo_2001, maass_real-time_2002}, whereby a recurrent neural network is initialized randomly and kept fixed, while only a linear output layer is trained. Recent neuroscientific evidence supports evidence for \gls{rc} properties in the  human cortex \citep{enel_reservoir_2016, buzsaki_space_2018}. As fixed weights do not suffer from forgetting, \citet{cossu_continual_2021-1} and \citet{kobayashi_continual_2019} applied \gls{rc} to \gls{cl}. However, in contrast to our method, both works are \emph{i)} on discriminative tasks and \emph{ii)} require task labels.

\subsection{Sequential Generative Continual Learning}
Although neuroscientific evidence shows that biological \gls{cl} in the human memory is based on generating neuronal sequences, most research on \gls{cl} focuses on discriminative tasks such as classification on static data (e.g., images). While this work applies \gls{cl} to \emph{generative} tasks on \emph{sequences}, past work has investigated the application of \gls{cl} for both 1) \emph{discriminative} tasks on \emph{sequences} \citep{sodhani_toward_2020, duncker_organizing_2020,cossu_continual_2021}, (see \citep{cossu_continual_2021, ehret_continual_2021} for reviews) and 2) \emph{generative} tasks on \emph{static} data using generative adversarial networks \citep{seff_continual_2017} or variational autoencoders \citep{nguyen_variational_2018}. Currently, the space for \emph{generative sequential} \gls{cl} is still a niche occupied by neuroscientifically inspired methods, forgoing standard backpropagation \citep{ororbia_continual_2020, ororbia_continual_2021}, or employing dendrite-inspired complex activation functions for \gls{cl} \citep{cui_continuous_2016,iyer_avoiding_2022,grewal_going_2021}. Although, like us, these works consider \emph{generative} tasks on \emph{sequences}, our work is novel as \emph{i)} we apply our method to \gls{ds} and \emph{ii)} we propse a \gls{cl} framework based on \gls{rc}.
\section{Conclusion}

    We propose a novel framework based on reservoir computing to continually learn dynamical systems from multiple environments. We compare favorably against baselines, showing promising results on several benchmark dynamical systems.
    With this study on reservoir computing for continual learning, we hope to inspire further research into:\begin{enumerate}[label=\emph{\roman*).}]
    	\item Sequential generative continual learning, which is currently overshadowed by the easier-to-quantify classification of static images.
	\item Reservoir computing as a training scheme for continual learning of sequences.
    \end{enumerate}
    We consider these two strands of research understudied and critically relevant in light of neuroscientific insights into human memory.

\section{Ethical Concerns}
    A brain-like continually learning memory is a crucial step towards general artificial intelligence (AI). While general AI can have unimaginably positive consequences, it also poses an existential risk to humanity. Apart from these general concerns, we do not foresee direct adverse effects of the presented research.
    On a positive note, a continually adapting learning algorithm for dynamical systems can be applied to all sciences that deal with complex spatiotemporal data and propel research on the frontier of high-dimensional nonlinear systems.

\newpage
\bibliography{main}
\bibliographystyle{collas/collas2022_conference}

\newpage
\appendix
\section{Appendix}

\subsection{Hyperparameter selection and evaluation}

\begin{figure}[!htp]
\centering
\begin{subfigure}{0.32\textwidth}
    \includegraphics[width=\textwidth]{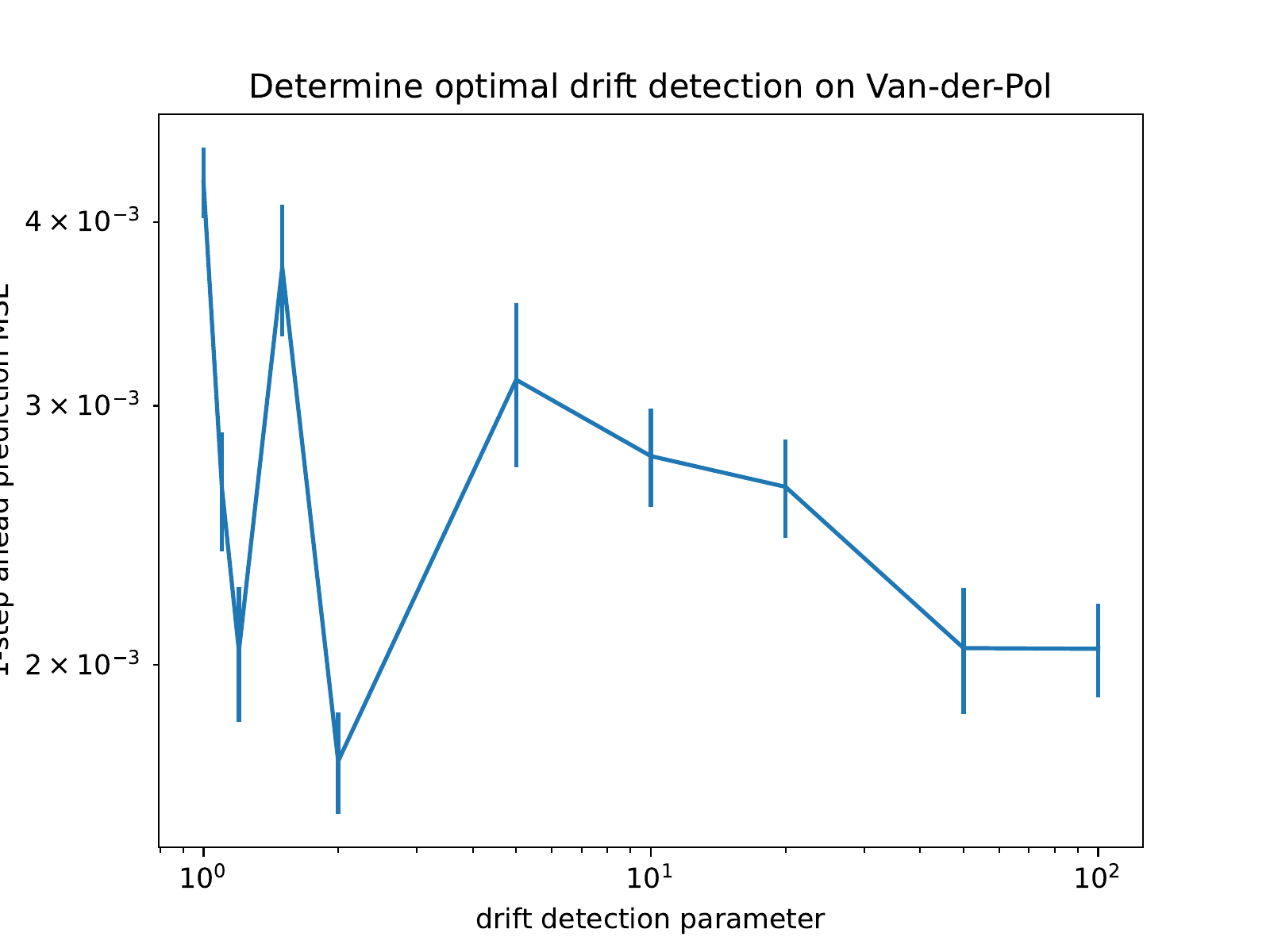}
    \caption{}
    \label{fig:thresh_vdp}
\end{subfigure}
\hfill
\begin{subfigure}{0.32\textwidth}
    \includegraphics[width=\textwidth]{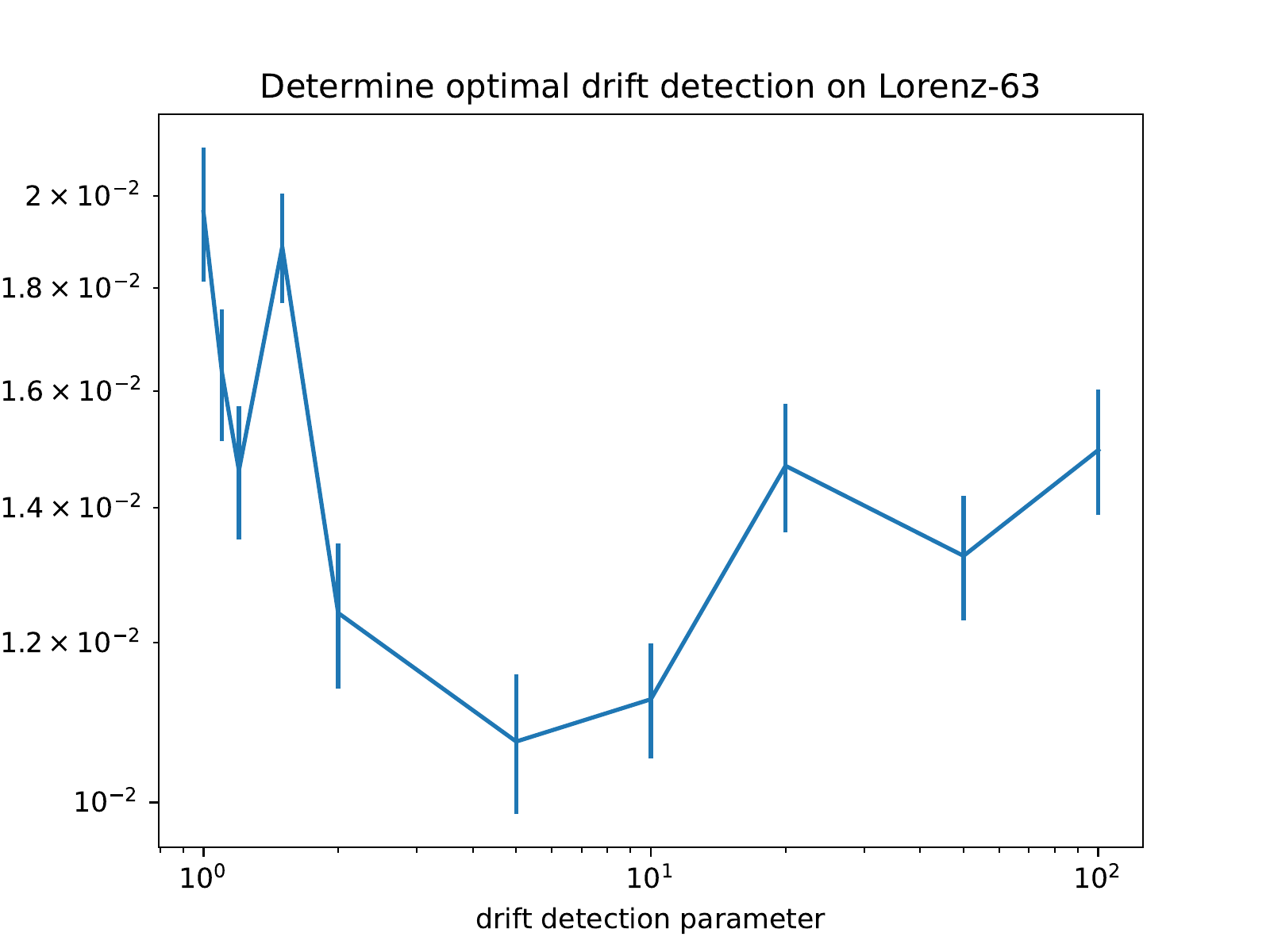}
    \caption{}
    \label{fig:thresh_l63}
\end{subfigure}
\hfill
\begin{subfigure}{0.32\textwidth}
    \includegraphics[width=\textwidth]{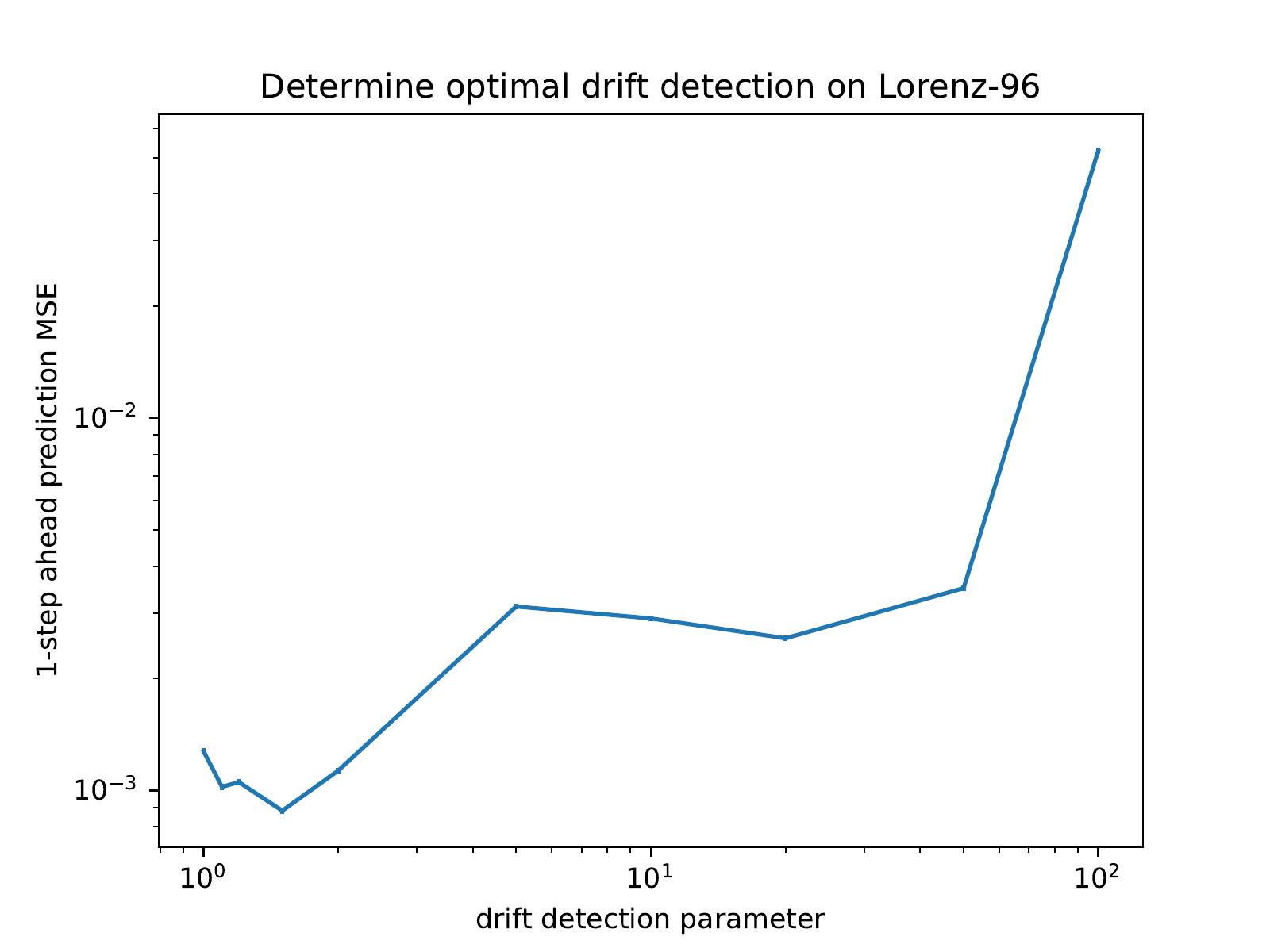}
    \caption{}
    \label{fig:thresh_l96}
\end{subfigure}

	\caption{Selecting the drift detection threshold for the (a) Van-der-Pol oscillator, the (b) Lorenz-63 attractor, and the (c) Lorenz-96 model.}
\label{fig:threshold}
\end{figure}

\begin{figure}[!htp]
\centering
\begin{subfigure}{0.32\textwidth}
    \includegraphics[width=\textwidth]{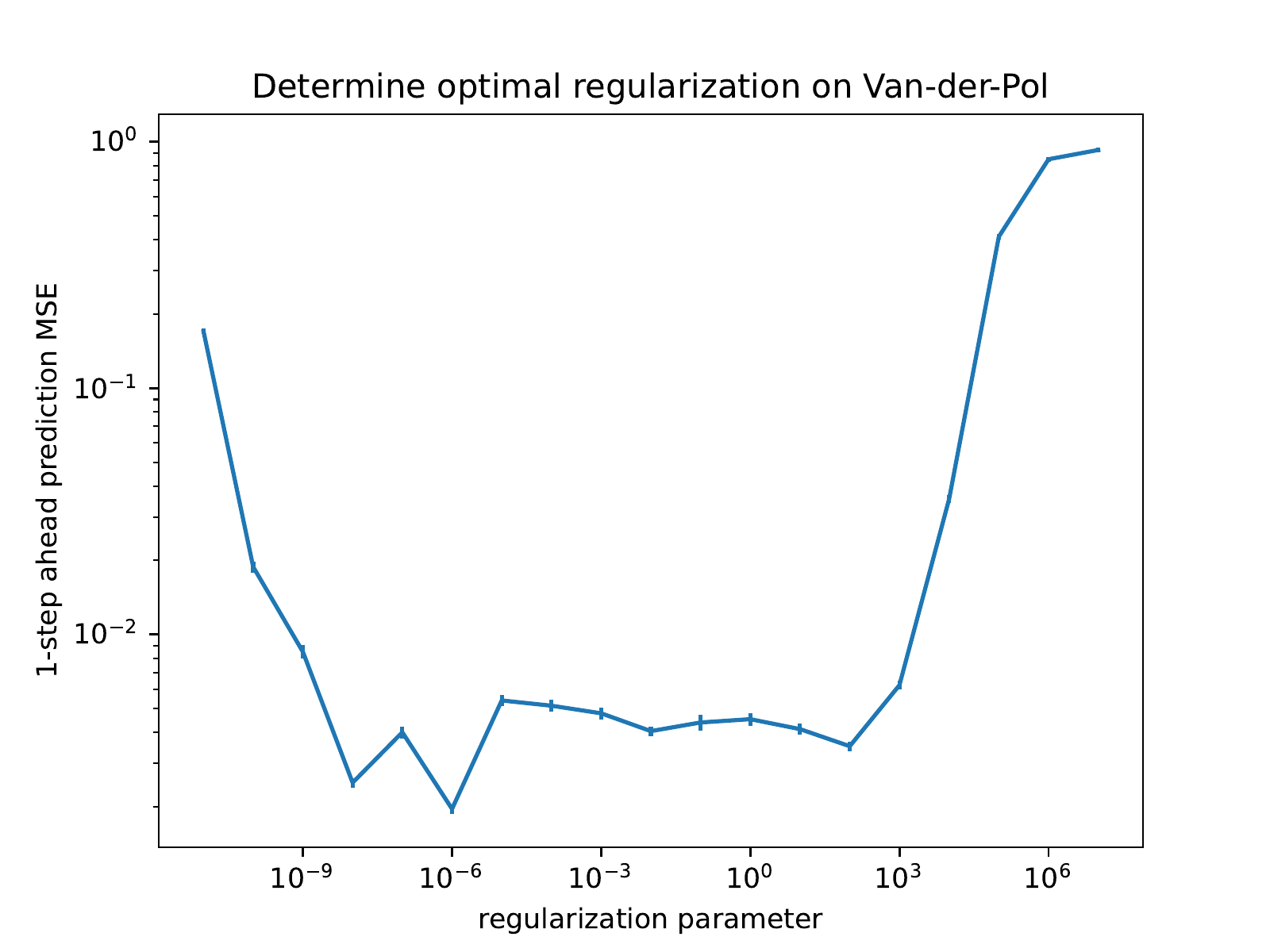}
    \caption{}
    \label{fig:reg_vdp}
\end{subfigure}
\hfill
\begin{subfigure}{0.32\textwidth}
    \includegraphics[width=\textwidth]{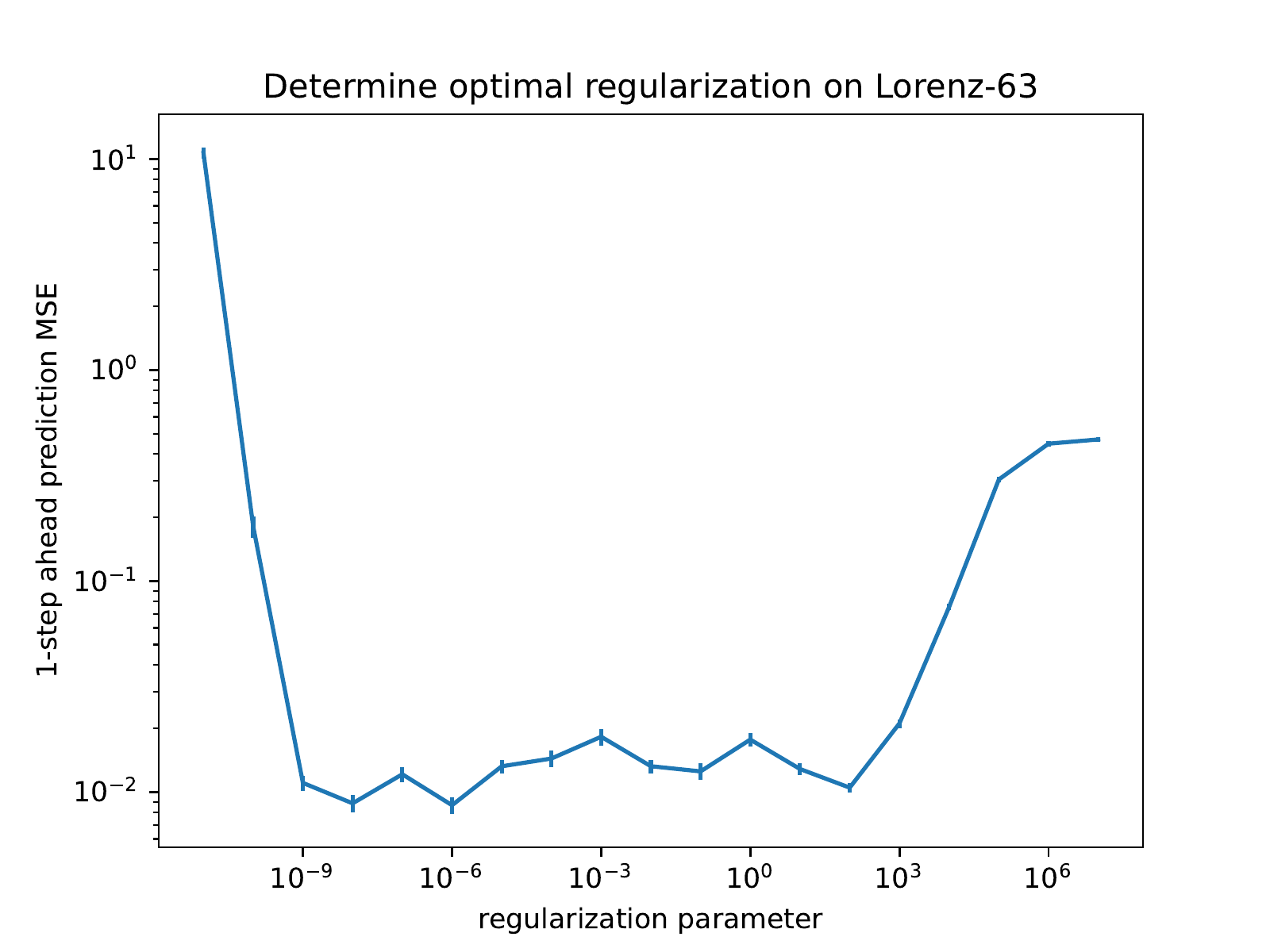}
    \caption{}
    \label{fig:reg_l63}
\end{subfigure}
\hfill
\begin{subfigure}{0.32\textwidth}
    \includegraphics[width=\textwidth]{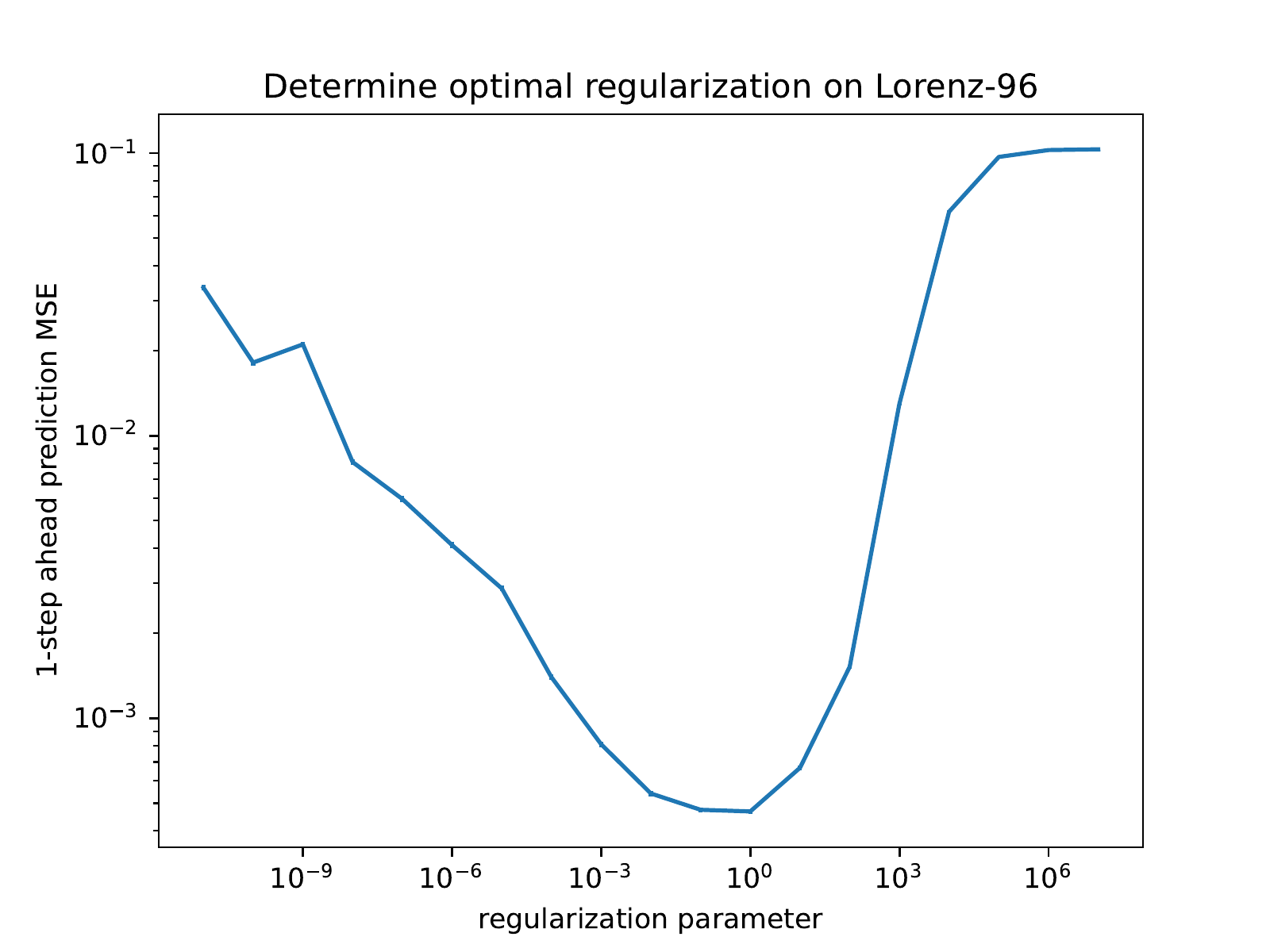}
    \caption{}
    \label{fig:reg_l96}
\end{subfigure}

	\caption{Selecting the regularization parameter for the (a) Van-der-Pol oscillator, the (b) Lorenz-63 attractor, and the (c) Lorenz-96 model.}
\label{fig:regularization}
\end{figure}

\begin{figure}[!htp]
\centering
\begin{subfigure}{0.32\textwidth}
    \includegraphics[width=\textwidth]{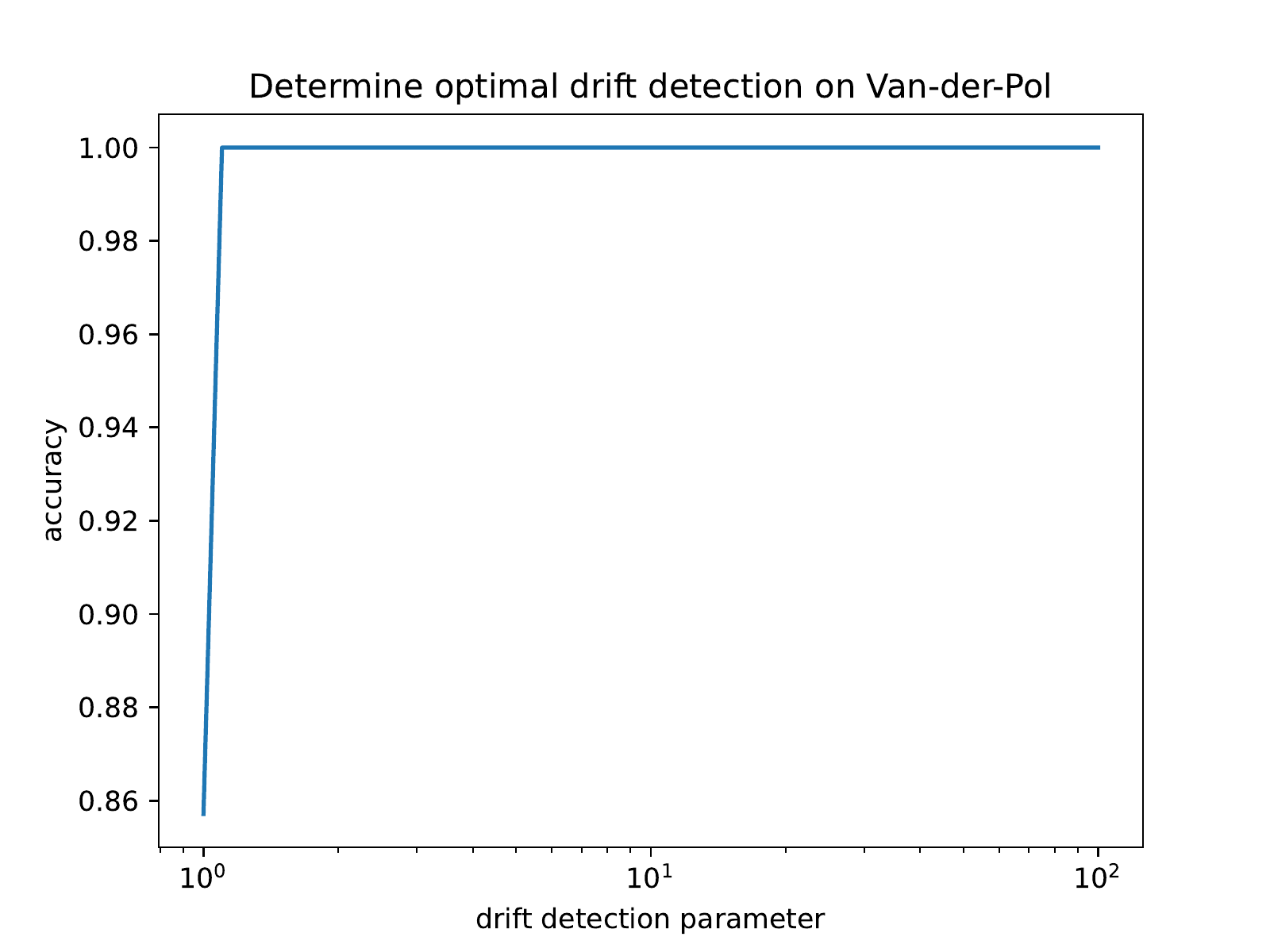}
    \caption{}
    \label{fig:accuracy_vdp}
\end{subfigure}
\hfill
\begin{subfigure}{0.32\textwidth}
    \includegraphics[width=\textwidth]{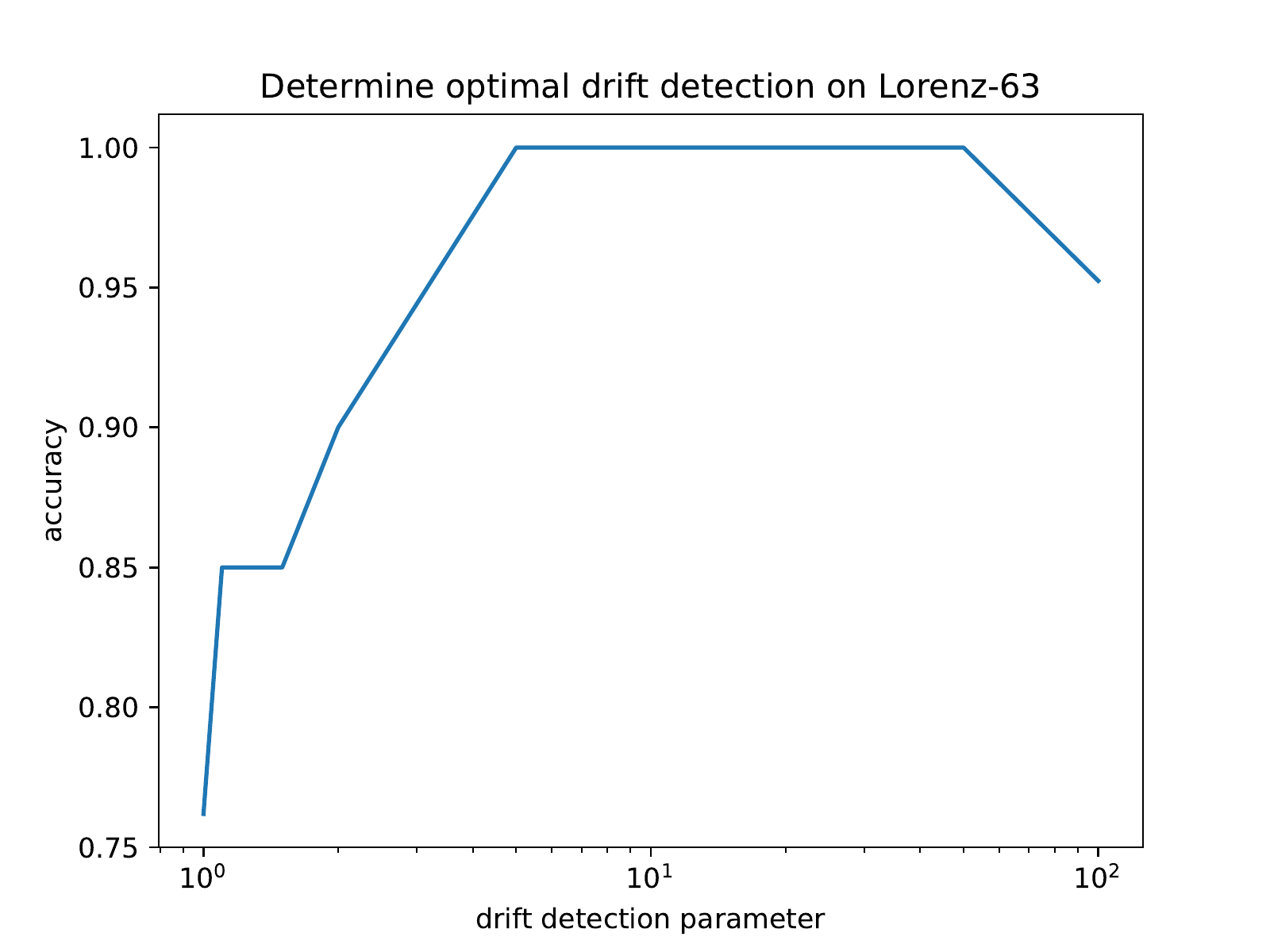}
    \caption{}
    \label{fig:accuracy_l63}
\end{subfigure}
\hfill
\begin{subfigure}{0.32\textwidth}
    \includegraphics[width=\textwidth]{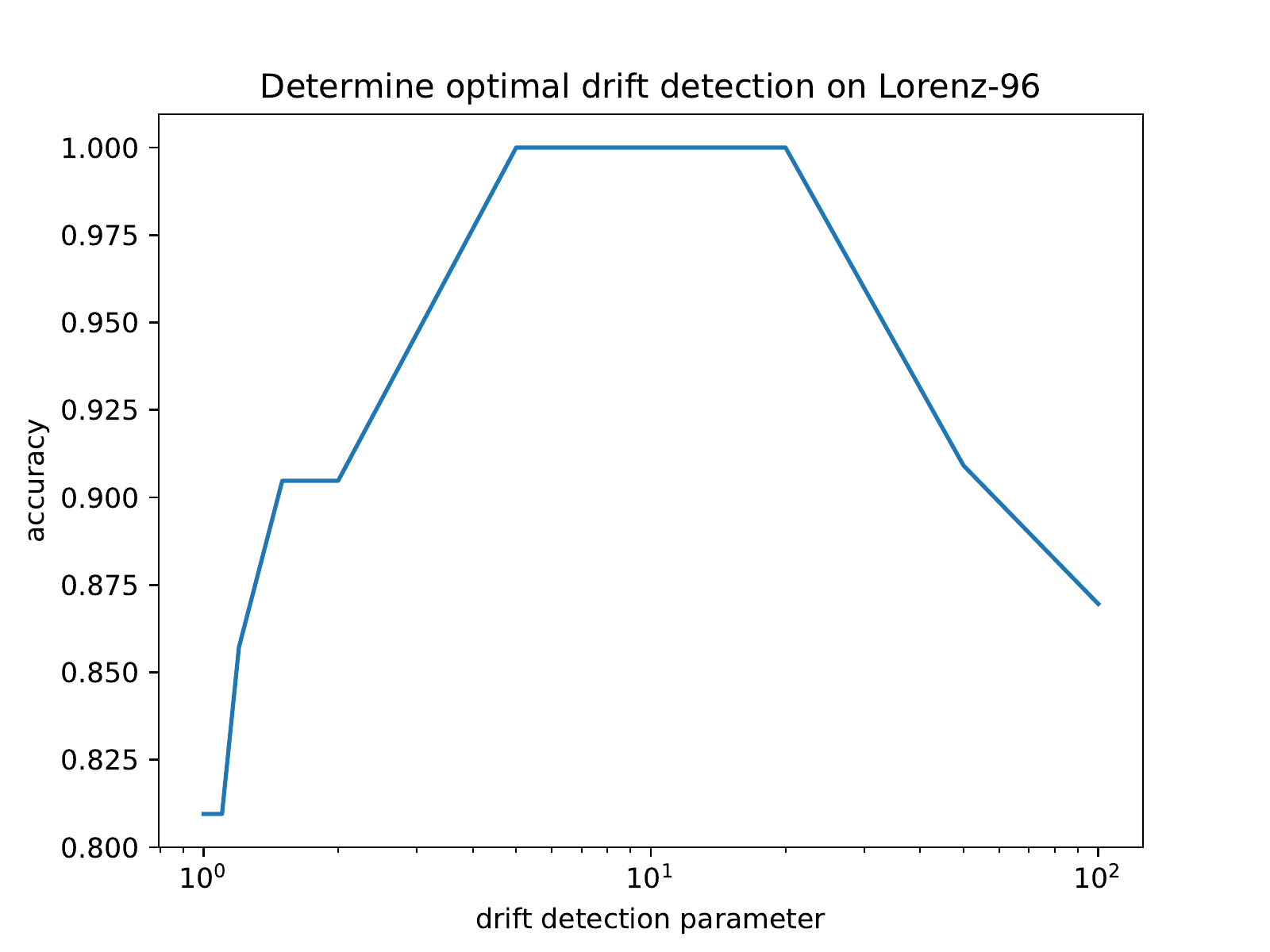}
    \caption{}
    \label{fig:accuracy_l96}
\end{subfigure}

	\caption{Evaluating the accuracy for different drift detection thresholds for the (a) Van-der-Pol oscillator, the (b) Lorenz-63 attractor, and the (c) Lorenz-96 model.}
\label{fig:accuracy}
\end{figure}

\newpage
\subsection{Performance comparison}

\begin{figure}[!htp]
\centering
\begin{subfigure}{0.32\textwidth}
    \includegraphics[width=\textwidth]{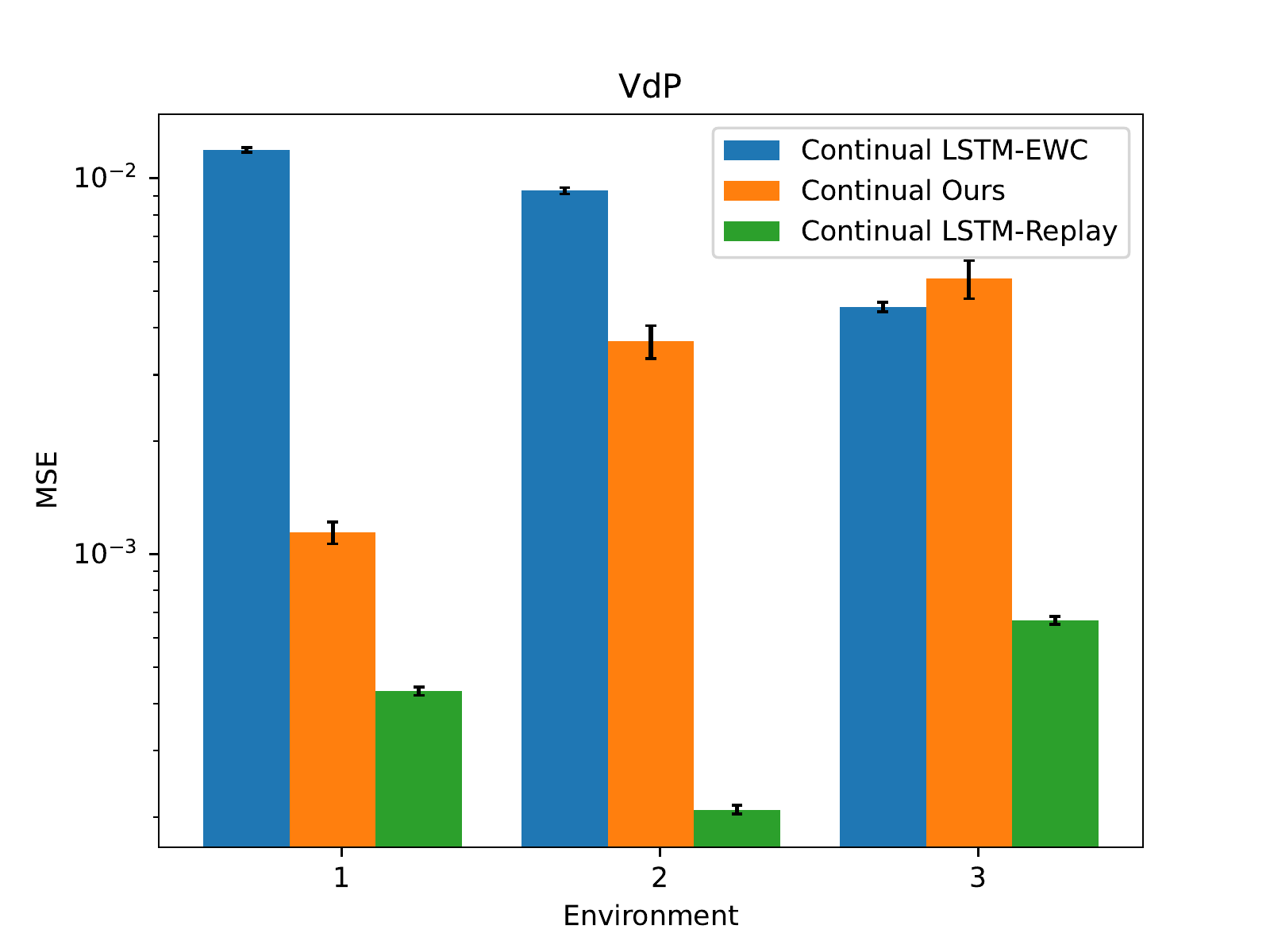}
    \caption{}
    \label{fig:continual_vdp}
\end{subfigure}
\hfill
\begin{subfigure}{0.32\textwidth}
    \includegraphics[width=\textwidth]{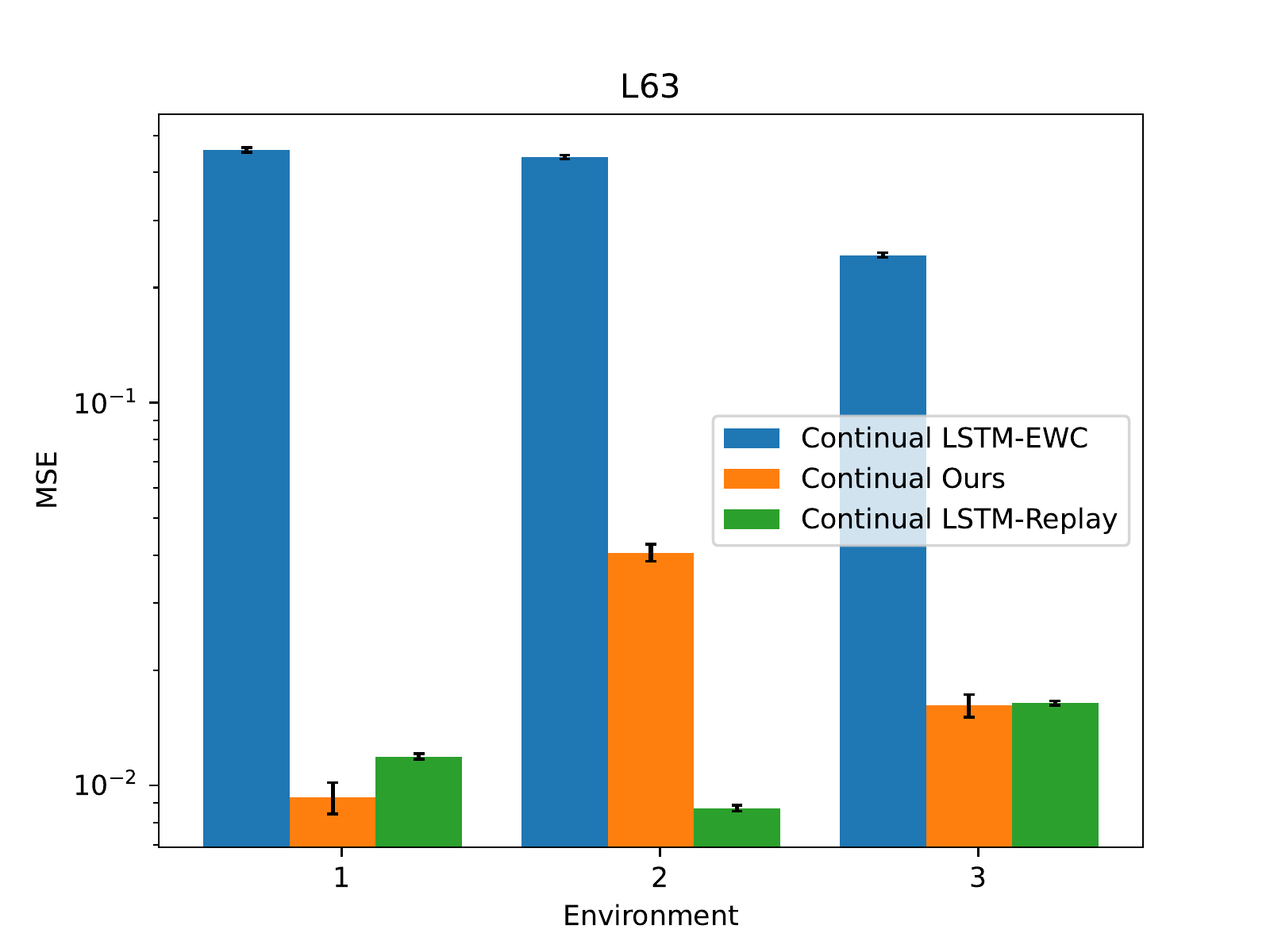}
    \caption{}
    \label{fig:continual_l63}
\end{subfigure}
\hfill
\begin{subfigure}{0.32\textwidth}
    \includegraphics[width=\textwidth]{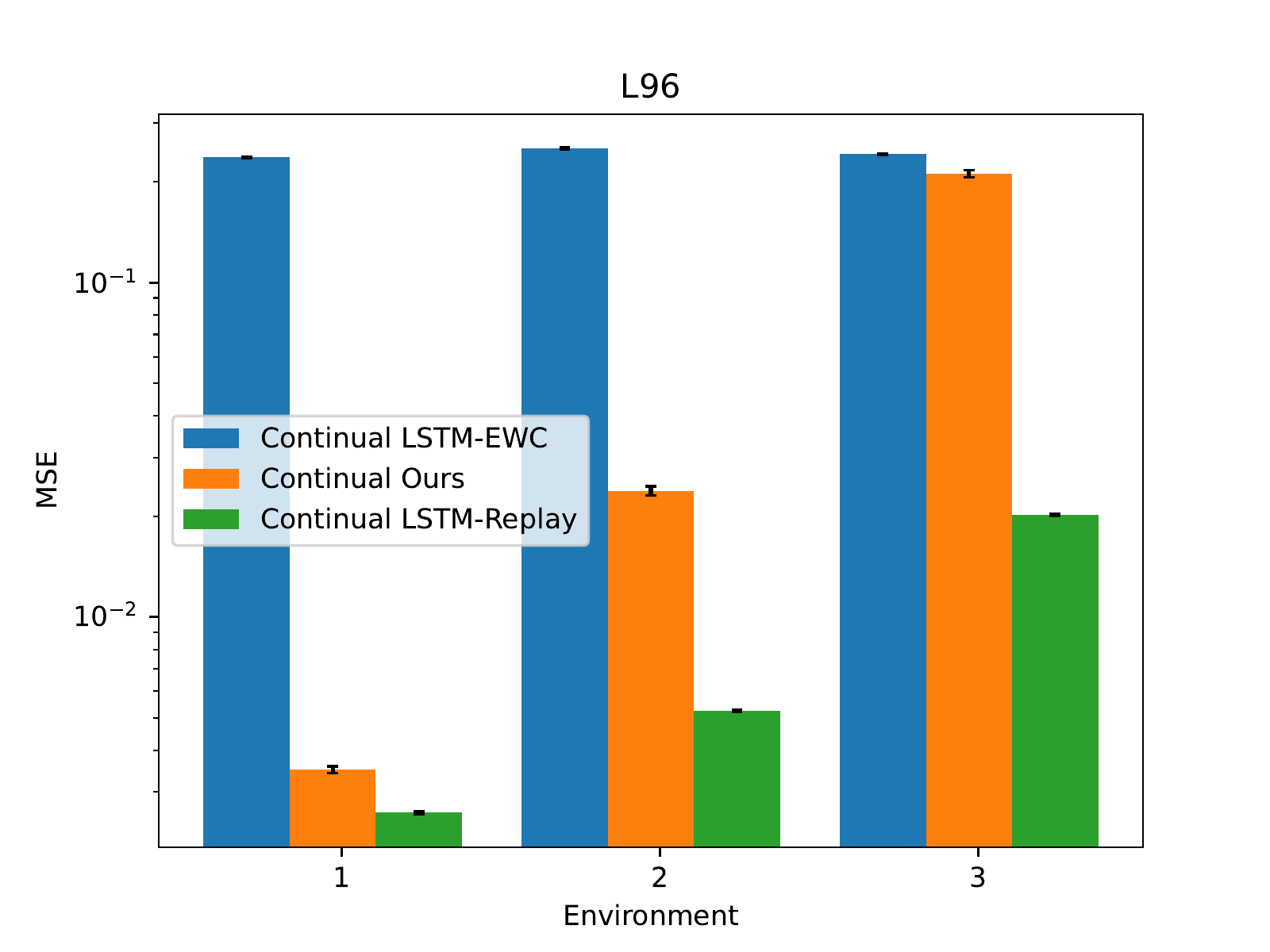}
    \caption{}
    \label{fig:continual_l96}
\end{subfigure}

	\caption{Comparing our method to continual learning baselines for the (a) Van-der-Pol oscillator, the (b) Lorenz-63 attractor, and the (c) Lorenz-96 model.}
\label{fig:continual}
\end{figure}

\begin{figure}[!htp]
\centering
\begin{subfigure}{0.32\textwidth}
    \includegraphics[width=\textwidth]{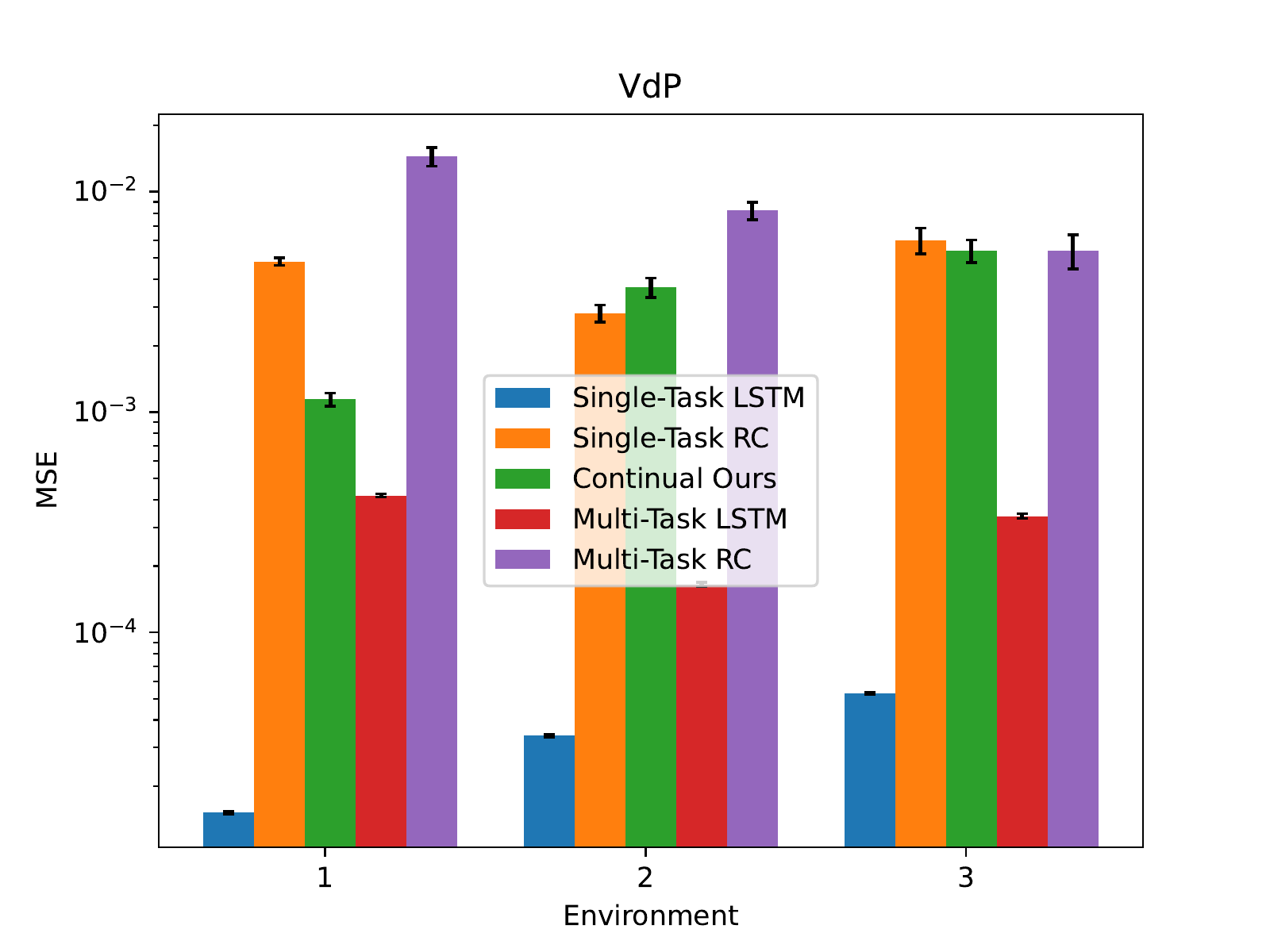}
    \caption{}
    \label{fig:bounds_vdp}
\end{subfigure}
\hfill
\begin{subfigure}{0.32\textwidth}
    \includegraphics[width=\textwidth]{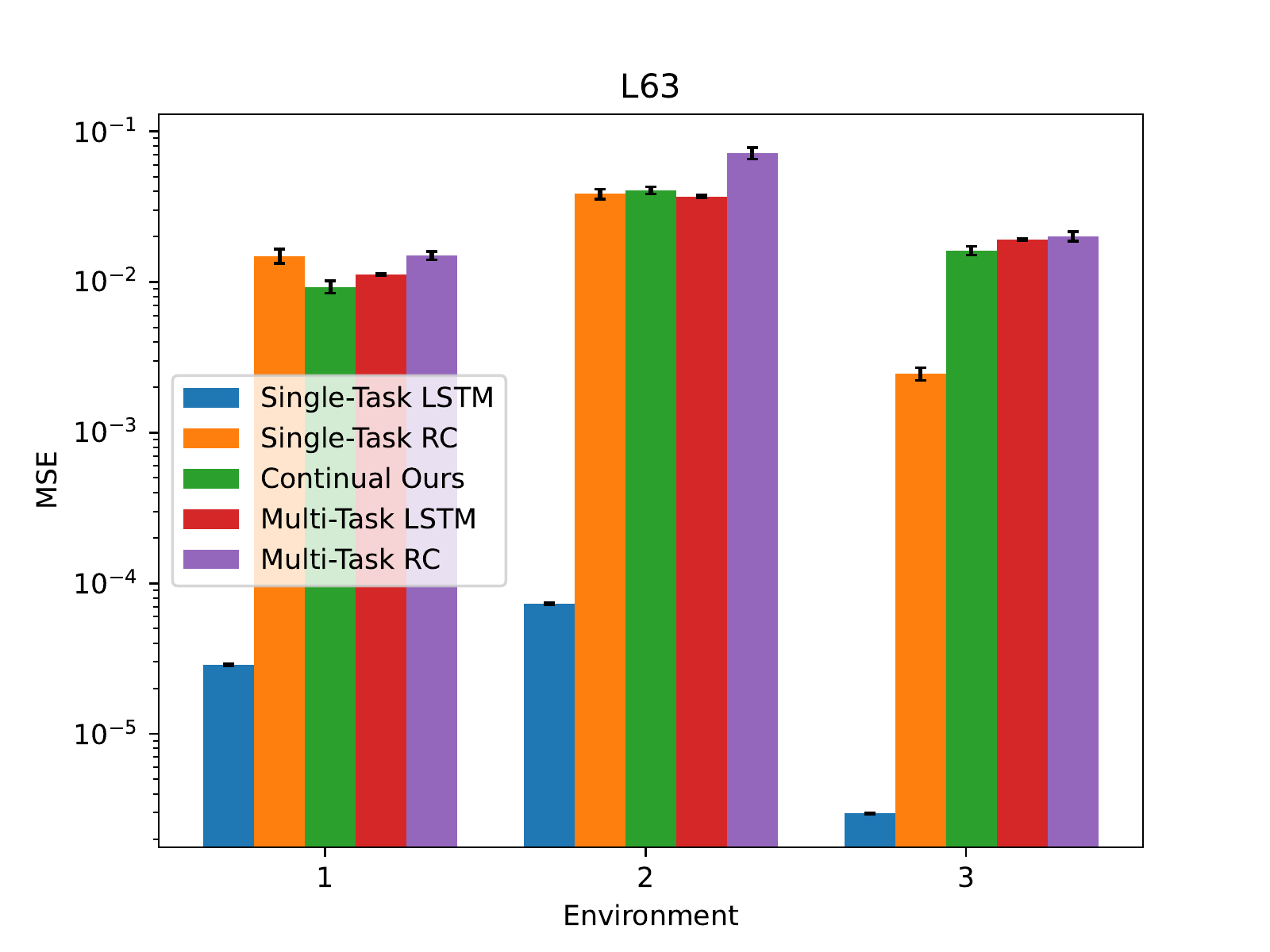}
    \caption{}
    \label{fig:bounds_l63}
\end{subfigure}
\hfill
\begin{subfigure}{0.32\textwidth}
    \includegraphics[width=\textwidth]{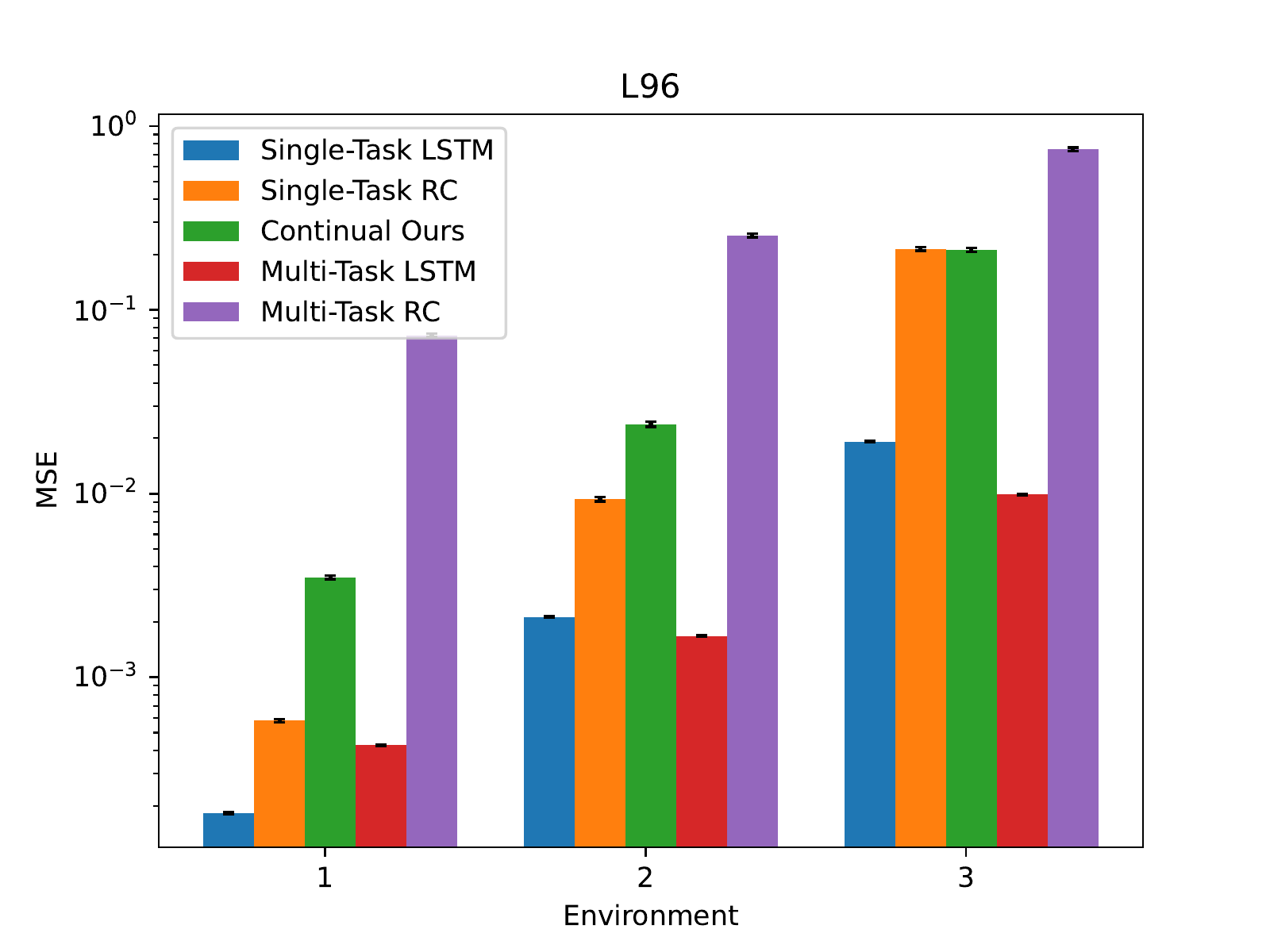}
    \caption{}
    \label{fig:bounds_l96}
\end{subfigure}

	\caption{Comparing our method to the single-task and multi-task settings for the (a) Van-der-Pol oscillator, the (b) Lorenz-63 attractor, and the (c) Lorenz-96 model.}
\label{fig:bounds}
\end{figure}

\subsection{Memory footprint}

Reservoir computing is not cheap on memory.
Given a reservoir size of $N_{\textrm{res}}$, the number of prediction heads $N_{\textrm{heads}}$, and the number of dataset dimensions $N_{\textrm{data}}$, the memory footprint of our model (to store the matrices $A$ and $B$) is $N_{\textrm{heads}} * N_{\textrm{res}} * N_{\textrm{res}} + N_{\textrm{heads}} * N_{\textrm{res}} * N_{\textrm{data}}$.
Thus, it grows linear with the number of prediction heads, similar to a simple replay baseline. It depends on the length and amount of sequences that are stored in the buffer which method would require more memory in terms of number of parameters.
However, we find the crucial difference here is that the parameters of our model do not store raw sensory data, while replay does.

\end{document}